\documentclass[10pt,conference,letterpaper]{IEEEtran}

\pdfoutput=1
\usepackage{amsmath,epsfig}
\usepackage{amsfonts}
\usepackage{times}
\usepackage{amssymb}
\usepackage{graphicx,subfig}
\usepackage{euscript}
\usepackage{algorithm, algorithmic}
\usepackage{color}
\usepackage{multirow}
\usepackage{booktabs}

\newcommand{\be}    {\begin{equation}}
\newcommand{\ee}    {\end{equation}}

\newcommand{{\hx}}  {\widehat{x}}

\title{A novel framework for image forgery localization}
\author{
{Davide Cozzolino, Diego Gragnaniello, Luisa Verdoliva}
\vspace{1.6mm} \\
\fontsize{10}{10}\selectfont\itshape
DIETI - University Federico II of Naples \\
Via Claudio 21, Naples - ITALY \\
\fontsize{9}{9}\selectfont\ttfamily\upshape
\{davide.cozzolino, diego.gragnaniello, verdoliv\}@unina.it
}

\begin{document}

\maketitle

\begin{abstract}
Image forgery localization is a very active and open research field
for the difficulty to handle the large variety of manipulations a malicious user can perform
by means of more and more sophisticated image editing tools.
Here, we propose a localization framework based on the fusion of three very different tools,
based, respectively, on sensor noise, patch-matching, and machine learning.
The binary masks provided by these tools are finally fused based on some suitable reliability indexes.
According to preliminary experiments on the training set,
the proposed framework provides often a very good localization accuracy and sometimes valuable clues for visual scrutiny.
\end{abstract}

\IEEEpeerreviewmaketitle

\section{Introduction}
This paper describes the strategy followed by the the GRIP team of the University Federico II of Naples (Italy)
to tackle the first IEEE IFS-TC Image Forensics Challenge on image forgery localization.

In order to deal with forgeries of different nature
(copy-paste from the same or a different image, exemplar-based inpainting)
we use techniques based on Photo Response Non Uniformity (PRNU) noise,
which deal uniformly with all these attacks,
and on which this team has gathered a solid experience \cite{CPPSV_wmfsi.10,CPPSV_dsp.11,CPSV_mmsp.13,CPSV_TIFS.13}.
This approach, however, relies on some strong hypotheses, not always satisfied.
In fact, it requires the knowledge of the camera PRNU itself, or else a sufficient number of images (at least a few dozens) to estimate it.
Therefore,
we decided to complement the PRNU-based technique with two more techniques oriented, respectively, to copy-move and splicing forgeries,
implementing eventually a simple decision fusion rule.
Specifically,
to localize copy-move forgeries we developed a simple technique based on the PatchMatch algorithm \cite{BSFG_TG.09} for fast block matching,
while to localize splicings we propose here a a new algorithm, based on some recently proposed local descriptors \cite{FK_TIFS.12}.

In the following of the paper we describe in detail the proposed strategy,
devoting Section II, III, and IV to forgery localization based on PRNU noise, PatchMatch, and local descriptors, respectively.
Section V describes the fusion algorithm and shows some results obtained on the test set.

\section{PRNU-based localization}

The PRNU pattern, originated by imperfections in the sensor silicon wafer,
is unique for each camera and stable in time, representing therefore a sort of a camera fingerprint,
which is present in all pristine images produced by the camera but absent in tampered areas.
By detecting the presence/absence of the camera PRNU in the image under test,
one is able to make reliable decisions on the presence of forgeries.

Let $y$ be a digital image observed at the camera output,
either as a single color band or the composition of multiple color bands.
with $y_i$ the value of pixel $i$.
In a simplified model \cite{CFGL_TIFS.08}, we can write $y$ as
\be
    y = (1+k)x + \theta = xk + x + \theta
\ee
where $x$ is the ideal noise-free image, $k$ the camera PRNU, $\theta$ an additive noise term which accounts for all types of disturbances, and products between images are pixel-wise.
Part of the ``noise'' can be removed by subtracting from $y$ an estimate of the true image, $\hx = f(y)$, provided by a denoising filter
obtaining the so-called noise residual, expressed after some manipulations as
\be
    r = y-\hx = yk + n = z + n
\ee
where all disturbances, including the denoising error, have been included in a single zero-mean noise term $n$.

The noise residual can be used for camera identification.
In fact,
the correlation index between $r$ and a given PRNU, $h$,
is a random variable with zero mean whenever $h\neq k$, and with a mean significantly different from zero only when $h=k$,
pointing to the camera that generated the image.
The same approach can be used to detect image forgeries,
by computing a correlation index field pixel-by-pixel by sliding a window of suitable size on the image,
and carrying out a local decision test.
When the computed correlation index $\rho_i$ is smaller than expected, a tampering of the corresponding pixel $i$ is likely.

In the concise description above, the camera PRNU pattern was assumed to be already available,
but this is only true if we have a collection of images taken by the camera large enough to carry out a reliable estimate.
However,
this is not the case in this challenge,
since we are only given a large number of images, with no information on their origin.
More precisely,
$N=$1500 training images are available, 1050 of them pristine and 450 fake,
while the test set comprises 700 fake images.
In principle, each of these images could have been taken by a different camera, frustrating any attempt to use a PRNU-based strategy.
However,
we rely on the reasonable conjecture that the unknown number of cameras $M$ used to build the database is much smaller than $N$.
Our algorithm comprises the following steps, described in detail in the rest of the Section:
\begin{itemize}
\item   group the training images in $C+1$ clusters (one for left-overs), based on their noise residuals;
\item   estimate the PRNU for the $C$ valid clusters;
\item   associate each test image with one of the clusters;
\item   localize forgeries.
\end{itemize}

\subsection{Implemented method}

Our first problem is to cluster the images based on their noise residuals.
At the end of the process,
clusters formed by a sufficient number of images will allow us to estimate the corresponding camera PRNU and perform forgery detection.
To carry out the clustering we use the algorithm proposed in \cite{Bloy_PAMI.08}
which is a simplified version of the well-known pairwise nearest neighbor (PNN) algorithm.
In PNN, at the beginning each data vector $v_j$ is the center of a cluster with just one element, $w_j=1$.
Then, the two closest centers, say $v'$ and $v''$ are merged together, provided they are closer than a given threshold,
generating by weighted averaging a new center that replaces the existing ones, in formulas
\begin{eqnarray}
    v_{\rm new} & = & (w'v'+w''v'')/(w'+w'') \nonumber \\
    w_{\rm new} & = & w' + w''
\end{eqnarray}
By so doing, the number of centers decreases by one at a time,
and the process continues until all centers are farthest apart than the threshold,
providing the desired clustering.
Even fast versions of PNN, however, are computationally demanding, as distances among all couples of data vectors must be computed.
The algorithm proposed in \cite{Bloy_PAMI.08} introduces some modifications to reduce computation time,
like picking at random couples to be compared with the threshold,
or looking for all points of a cluster before proceeding with another one.

In our case, the data vectors are the normalized noise residuals $r_j/y_j=k_j+n_j/y_j$,
which represent basic estimates of the camera PRNU that are gradually improved through merging.
The distance measure is the Peak to Correlation Energy ratio (PCE) \cite{GF_wdw.08}, more robust than the correlation index.
We carry out the clustering on the training set using a threshold equal to 50.
By so doing we identify 44 different clusters, for a total of 746 pristine images out of the 1050 available and 315 fakes out of 450 (see Fig.1).

Although in the clustering phase we estimate the PRNU by unweighted averaging of the normalized noise residuals,
the final estimate for the cluster $C$ is computed as:
\be
    \widehat{k}_C = \sum_{j \in C} y_j r_j / \sum_{j \in C} y_j^2
\ee
where the weighting terms $y_j$ account for the fact that dark areas of the image present an attenuated PRNU
and hence should contribute less to the overall estimate.

At this point we can try to associate the test images with one of the estimated PRNU's using again PCE.
With a threshold equal to 100 we are able to classify 431 of the 700 images available, about 60\% of the total, shown in Fig.1.

\begin{figure}[t]
\centering
\begin{minipage}[c]{.99\linewidth} \centerline{\epsfig{figure=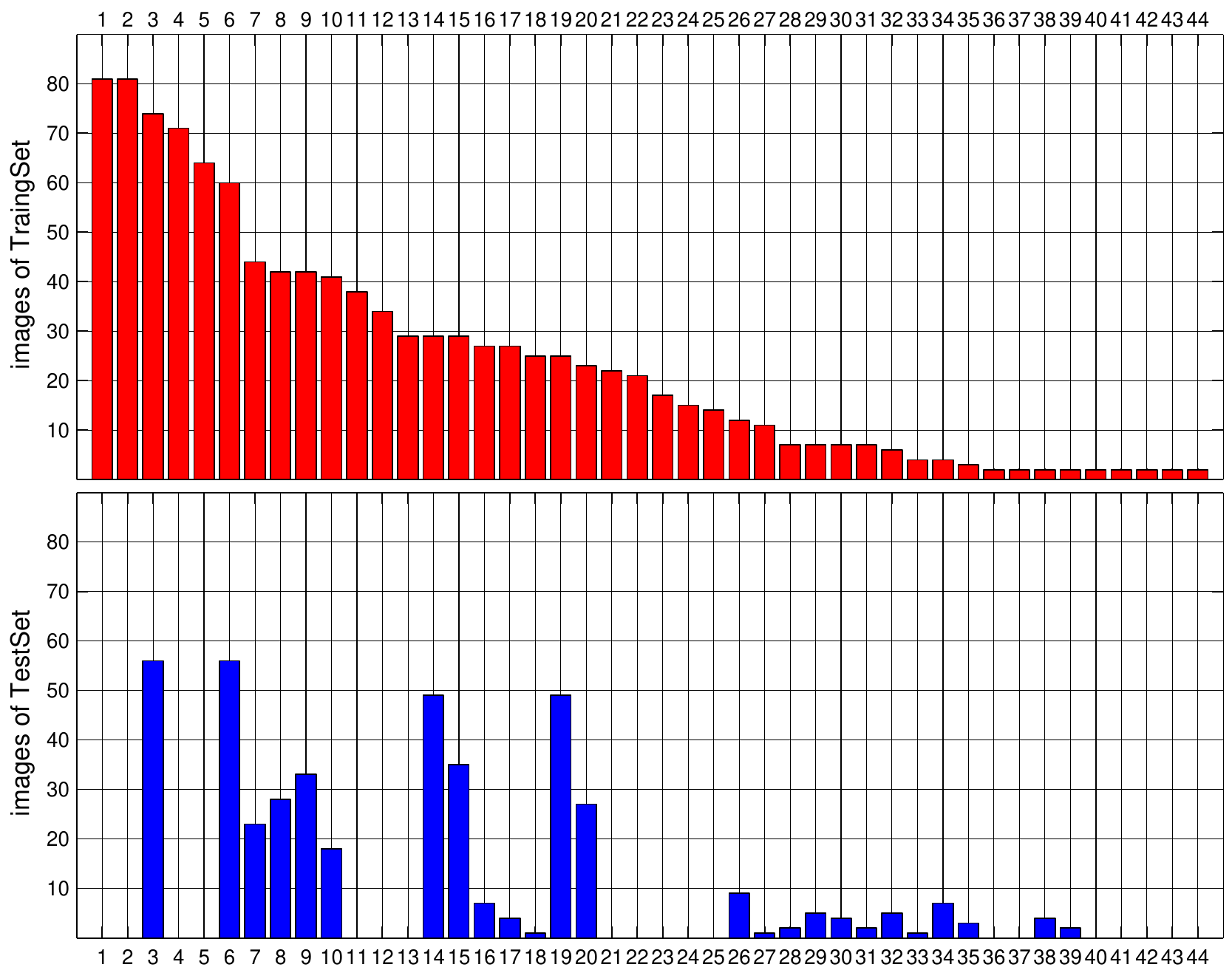, width=9cm}} \end{minipage}
\caption{Number of images belonging to the clustered sets.}
\label{fig:clusters}
\end{figure}

For all forged images belonging to one of the identified clusters,
forgery detection is carried out as proposed in \cite{CFGL_TIFS.08}
using the normalized correlation index between $r_{_{W_i}}$ and $z_{_{W_i}}$,
the restrictions of $r$ and $z$, respectively, to the 129$\times$ 129 window $W_i$ centered on the target pixel.
There are two main differences with respect to the original algorithm.
First, to improve the quality of the noise residuals we resort to nonlocal denoising.
This choice, as shown in \cite{CPPSV_wmfsi.10, CPSV_TIFS.13}, improves the separation between image content and PRNU, especially in textured areas.
In addition, we use an adaptive decision threshold here,
which depends on the reliability of the correlation field, measured through PCE.
In fact, given the lack of information on the camera used to take the photos, correlation fields are not equally reliable.

It is worth underlining that the correlation might happen to be very low when the image is dark, saturated or strongly textured,
increasing the false alarm probability in these areas.
In \cite{CFGL_TIFS.08} this problem is addresses by means of a ``predictor'' which,
based on local images features, such as texture, flatness and intensity,
computes the expected value of the correlation index under the hypothesis that PRNU is present.
In this work we do not use the predictor, as it proves unreliable when estimated only on a few images.
However, we keep enforcing a control on saturated areas, where PRNU is totally unreliable.
In Fig.2 we show two images of the training set
with the corresponding correlation maps (low values correspond to red in this case) and detection masks.

\setlength\fboxsep{0.8pt}\setlength\fboxrule{0.5pt}

\begin{figure}[t]
\centering
\includegraphics[width=0.15\textwidth]{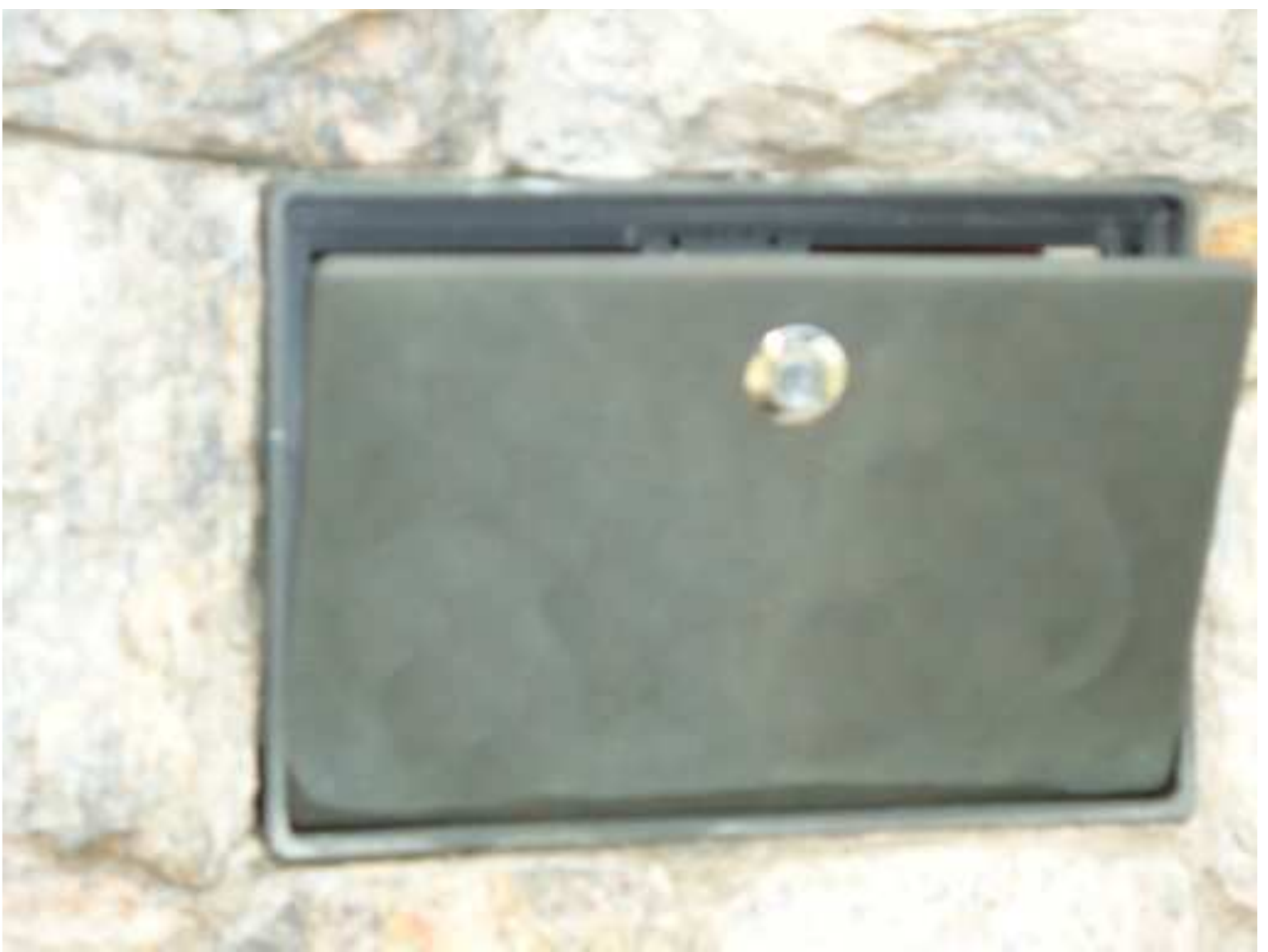}
\includegraphics[width=0.15\textwidth]{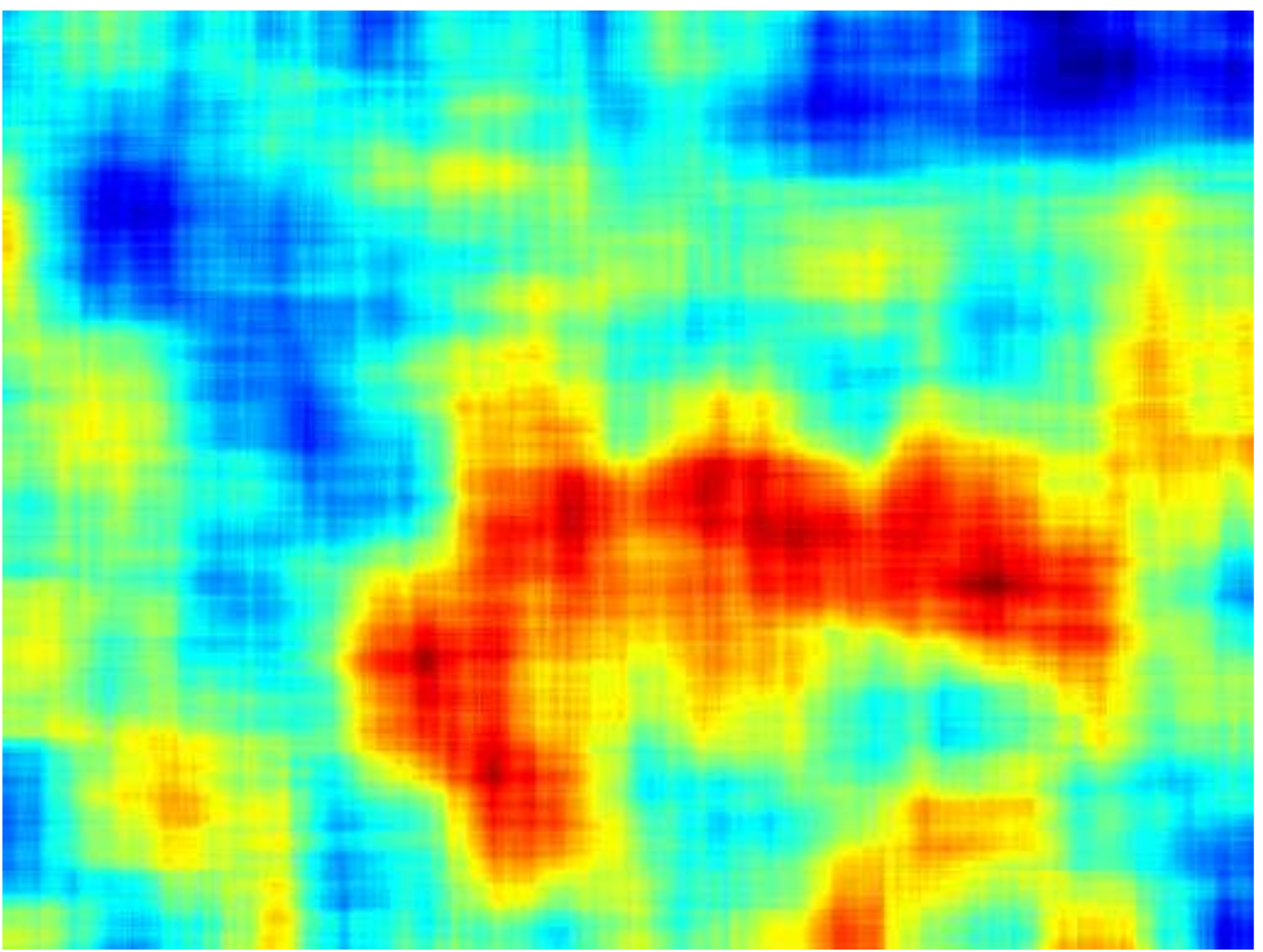}
\includegraphics[width=0.15\textwidth]{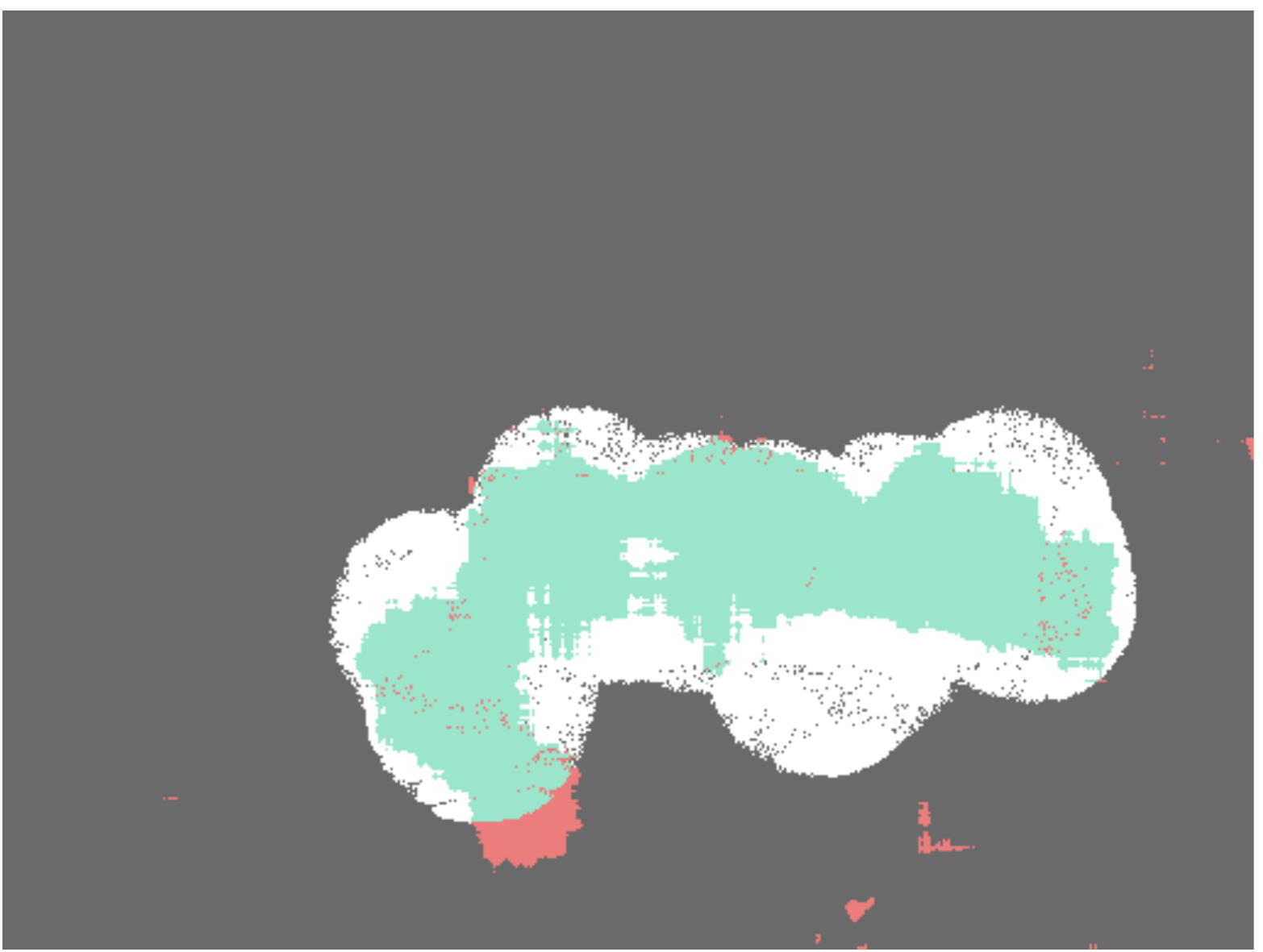}

\vspace{3mm}
\includegraphics[width=0.15\textwidth]{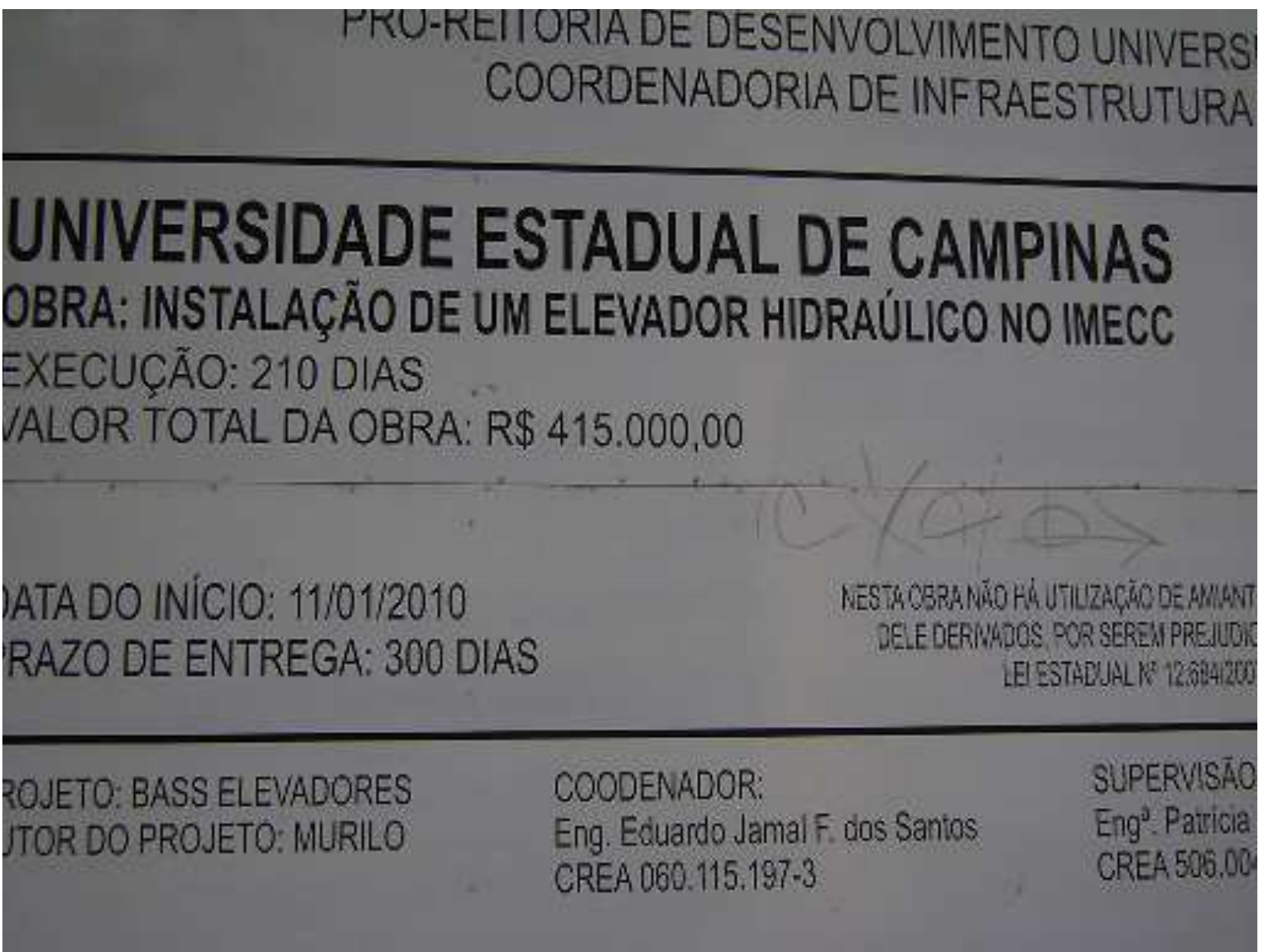}
\includegraphics[width=0.15\textwidth]{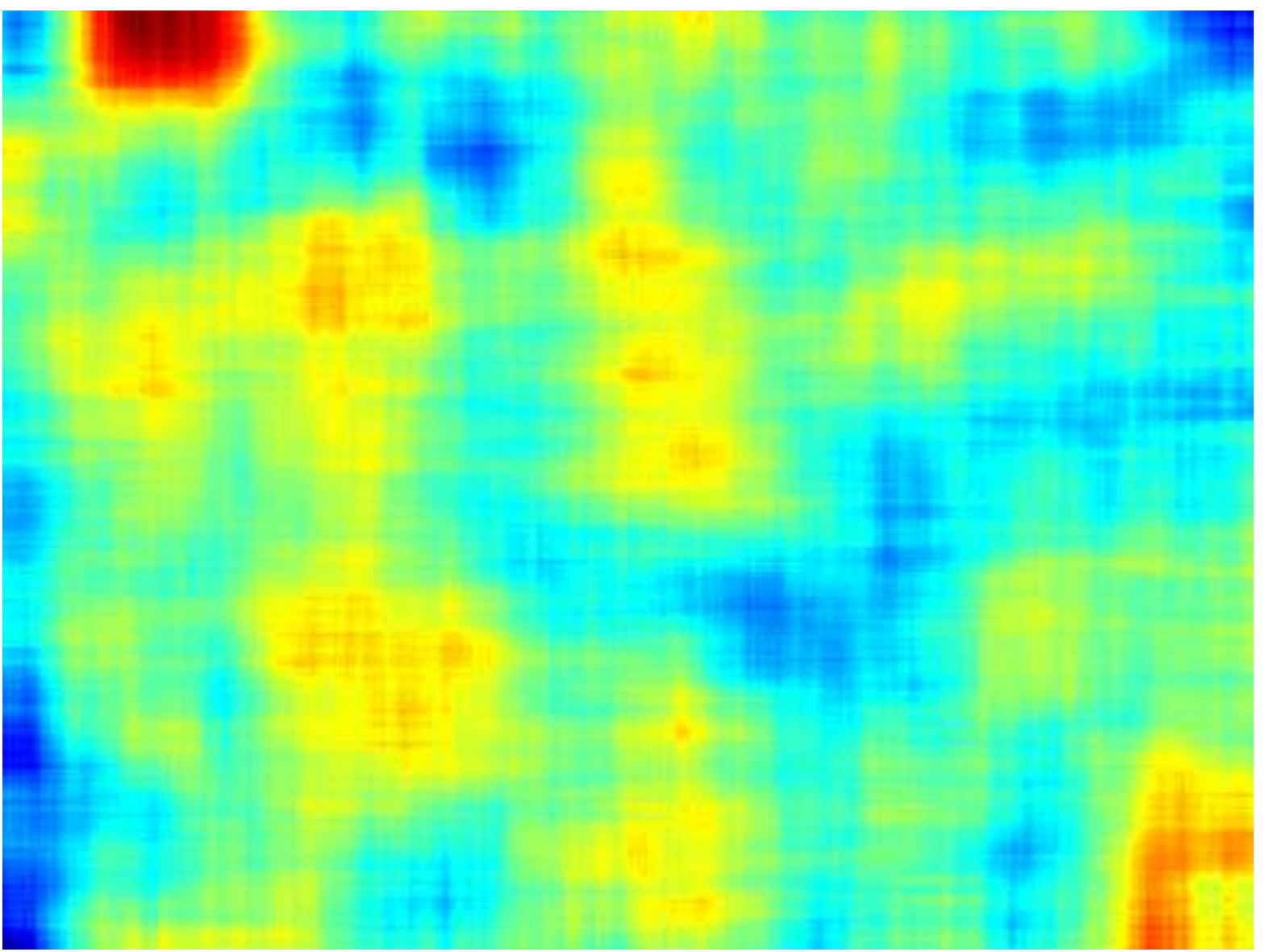}
\includegraphics[width=0.15\textwidth]{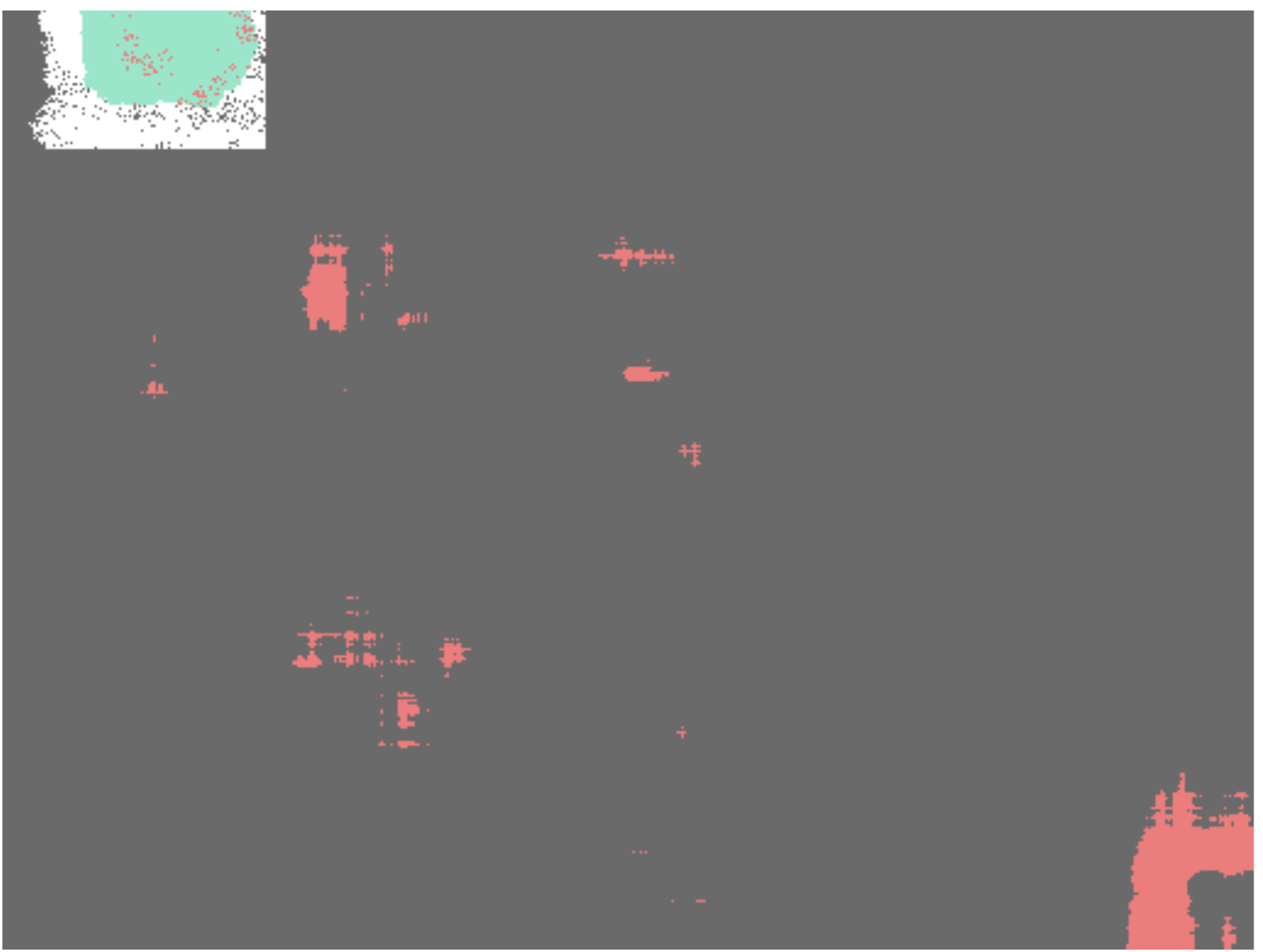}

\caption{Two training fake images, correlation maps and color-coded detection masks.
Gray: genuine pixel declared genuine,
red: genuine pixel declared tampered (error),
white: tampered pixel declared genuine (error),
green: tampered pixel declared tampered.} \label{fig:copymoves}
\end{figure}

\section{Copy-move forgery localization based on PatchMatch}

Localization of copy-move forgeries is a very active field of research and several papers face the problem,
the majority of which based on keypoint identification \cite{CRJA_TIFS.12} followed by feature extraction and matching.
This approach works quite well for classical copy-move forgeries,
where a large compact object is copied from source to target location, with some possible modification (rotation, resizing, and so on) \cite{PL_TIFS.10}.
Things are more difficult when multiple small regions are copied from all over the image
and combined together to cover a large object (exemplar-based inpainting),
since keypoint identification and feature matching becomes much quite unreliable.
In this case,
better results can be obtained by computing a dense motion field by some block-matching algorithm, as done in \cite{LG_crv.06, CYC_IVC.13}.

We have followed a similar line of work, resorting to PatchMatch \cite{BSFG_TG.09},
a recently proposed editing algorithm, which provides an accurate (though approximate) motion field much faster than exact algorithms.
The main steps of the localization algorithm are the same as in the methods based on feature extraction \cite{CRJA_TIFS.12},
namely, (dense) motion field estimation, filtering and post-processing.
In particular,
matching is performed directly on the RGB image, normalized to gain robustness against changes of illumination, with 7$\times$7-pixel patches.
Once the motion vector field is computed,
we single out regions with homogeneous motion by a suitable linear filtering (robust to moderate resizing).
To avoid false alarms we remove matches between spatially close regions, and matches obtained in perfectly flat areas, as in the presence of saturation.
Then for each motion vector we compare the image with its shifted version and compute a dense correlation map which,
after thresholding and morphological operations, provides the binary map relative to a single copied object.
Of course, we detect both the source and target regions, associated with opposite motion vectors.
To deal with rotations and relatively large resizing,
we evaluate the motion vector field for a fixed number of rotations and resizings, taking advantage of PatchMatch speed.

Two sample results on the training set are shown in Fig.3,
referring to a classical copy-move and an exemplar-based inpainting.
Note that the algorithm is not able to distinguish the original object from the copy.
However,
we can use the information coming from the PRNU-based approach (when available) on remove this uncertainty as in the example of Fig.4.
This technique, however, is reliable only when the tested objects are relatively large
and the correlation map is sufficiently reliable (PCE$>$150),
in all other cases we declare both regions as forged.

\begin{figure}[t]
\centering
\includegraphics[width=0.15\textwidth]{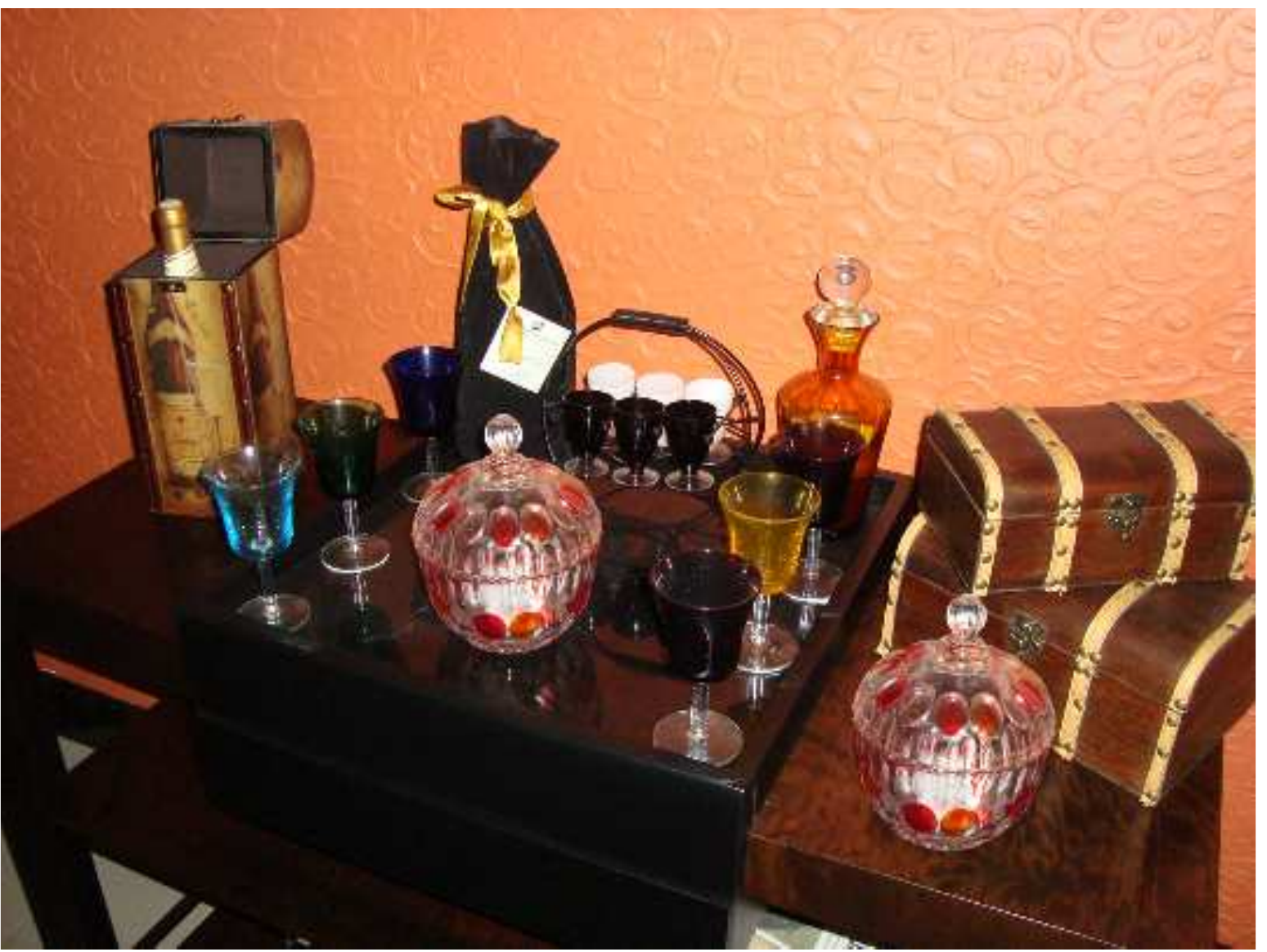}
\fbox{\includegraphics[width=0.15\textwidth]{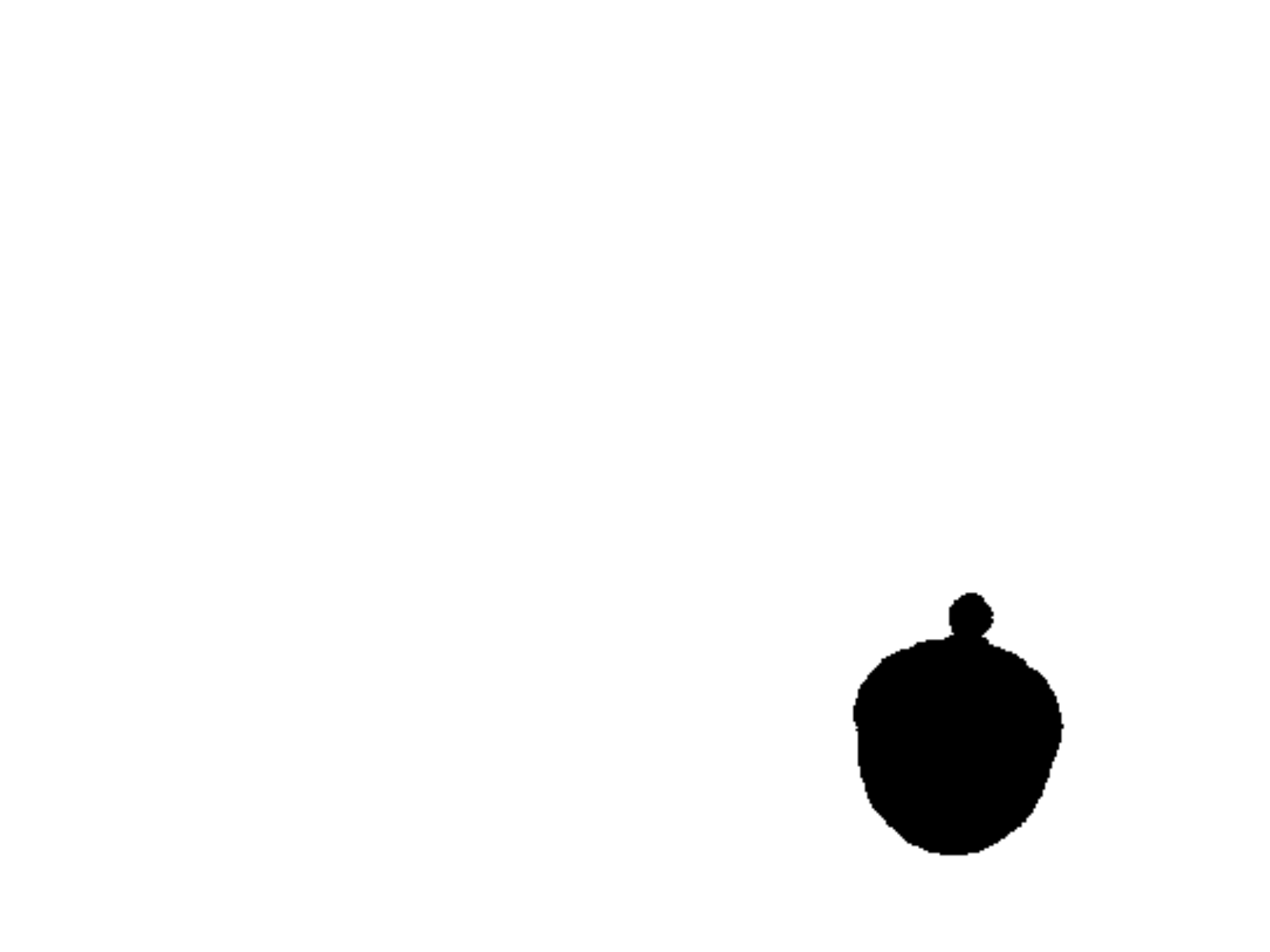}}
\fbox{\includegraphics[width=0.15\textwidth]{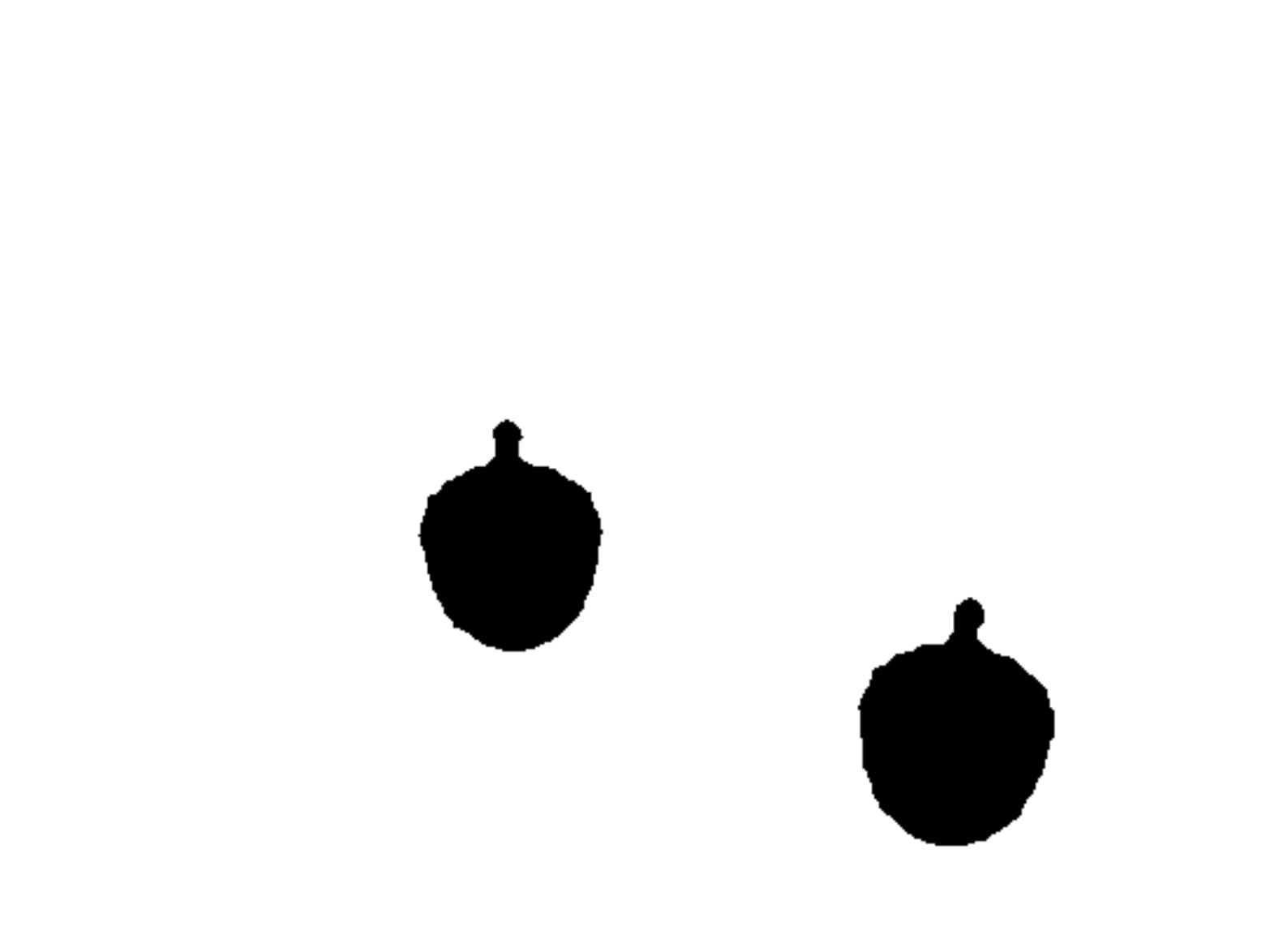}}

\vspace{3mm}
\includegraphics[width=0.15\textwidth]{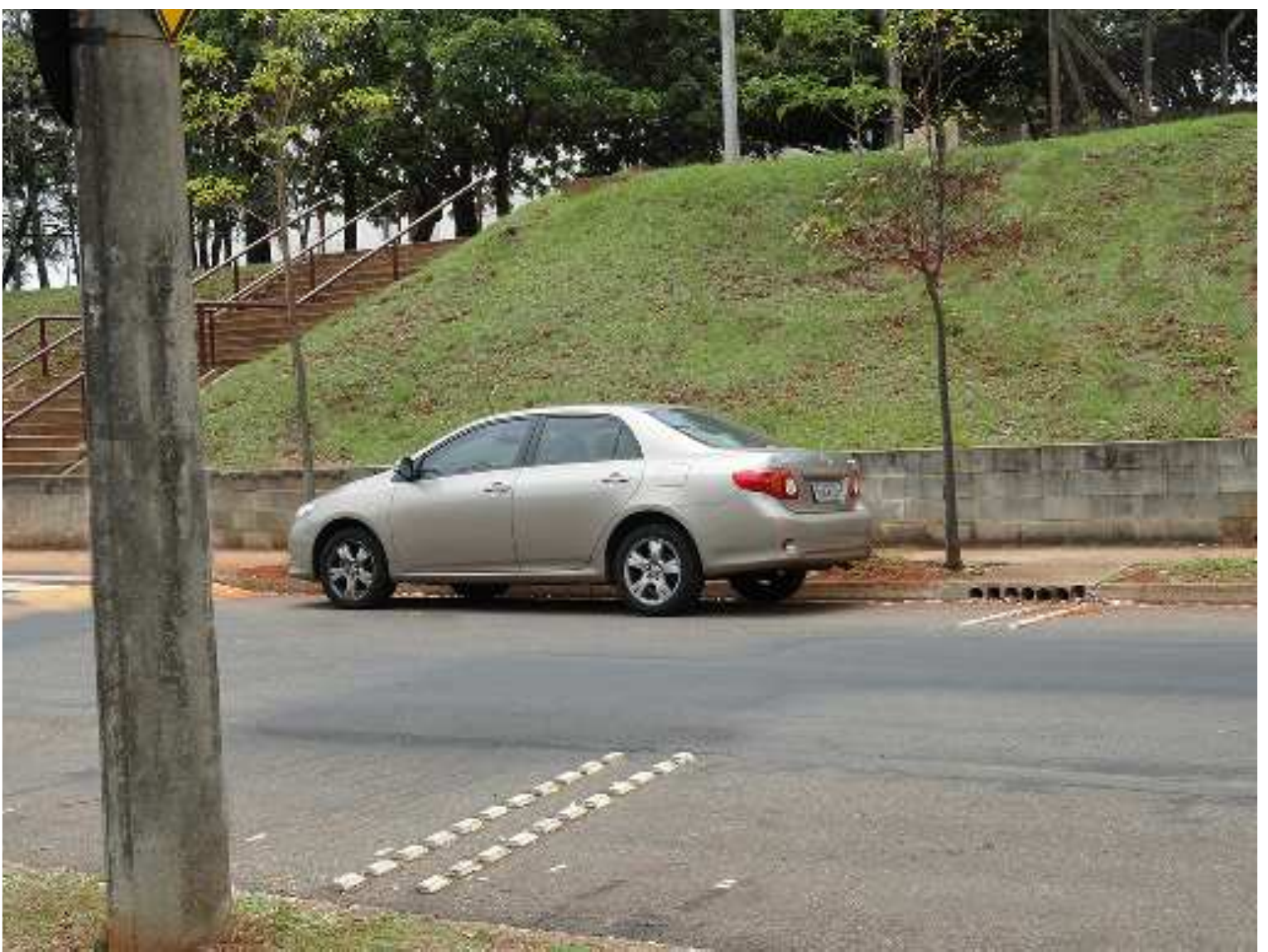}
\fbox{\includegraphics[width=0.15\textwidth]{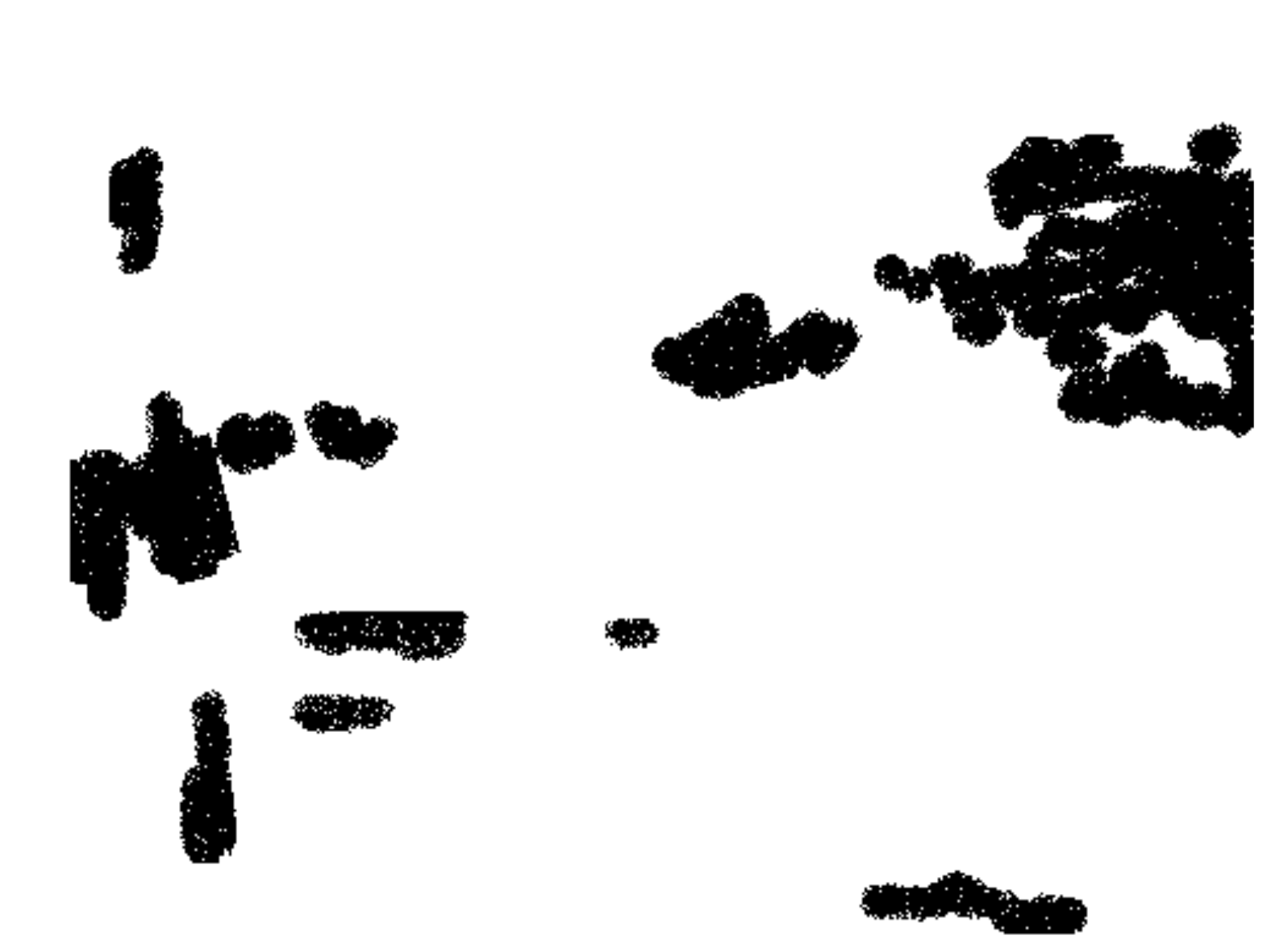}}
\fbox{\includegraphics[width=0.15\textwidth]{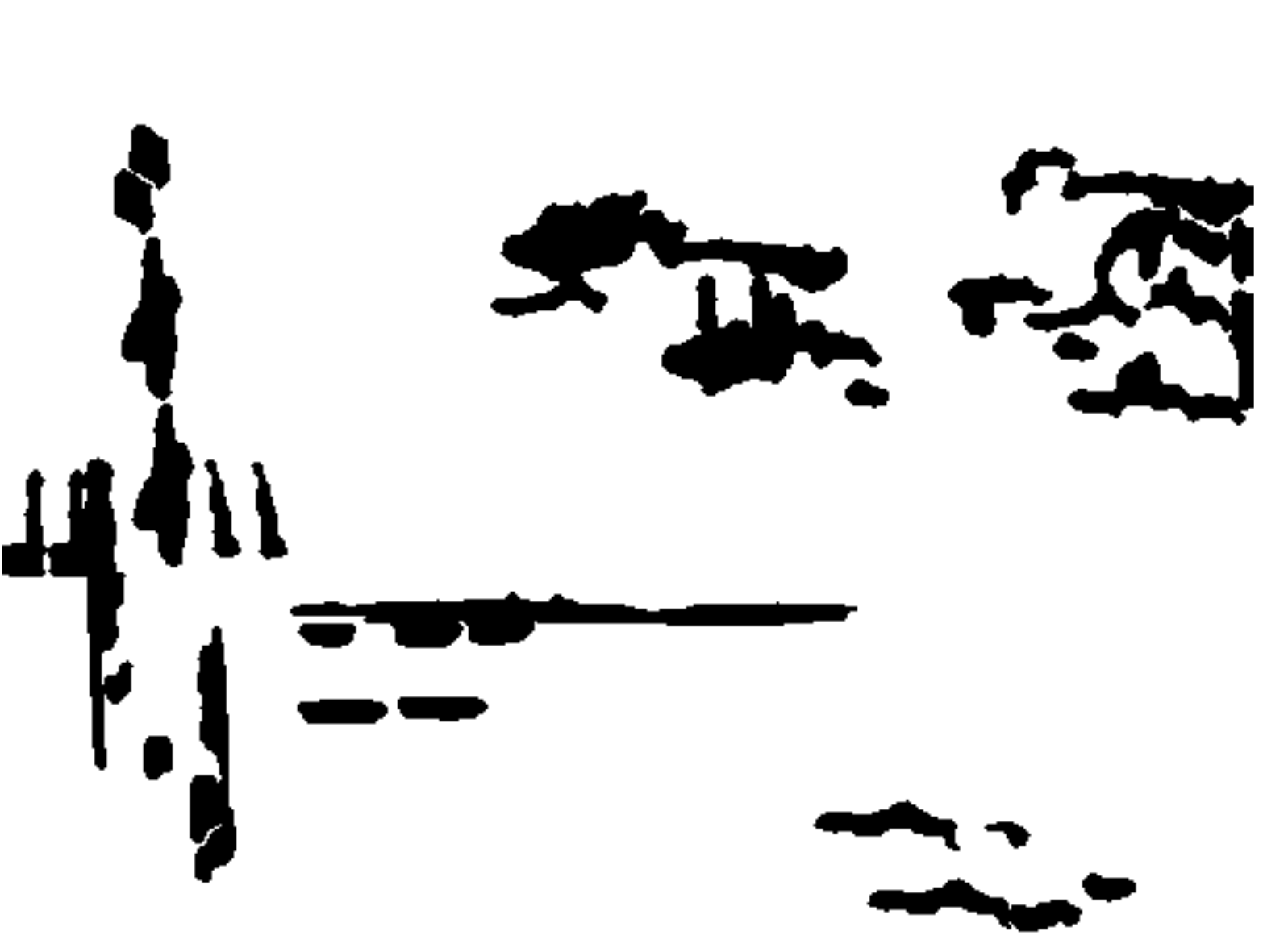}}
\caption{Two training fake images, their ground truth, and the output of our algorithm.} \label{fig:copymoves}
\end{figure}

\begin{figure}[b]
\centering
\includegraphics[width=0.11\textwidth]{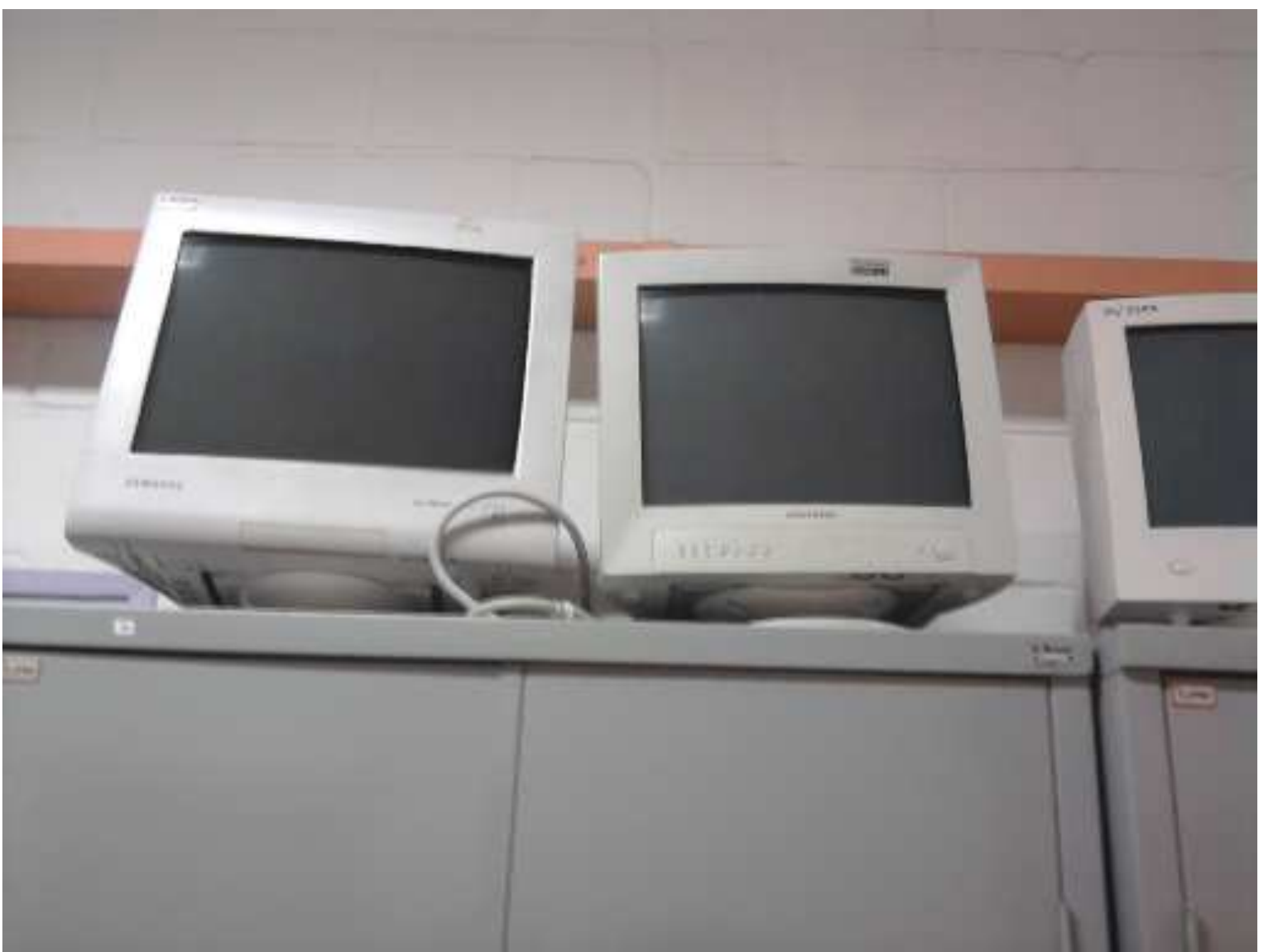}
\includegraphics[width=0.11\textwidth]{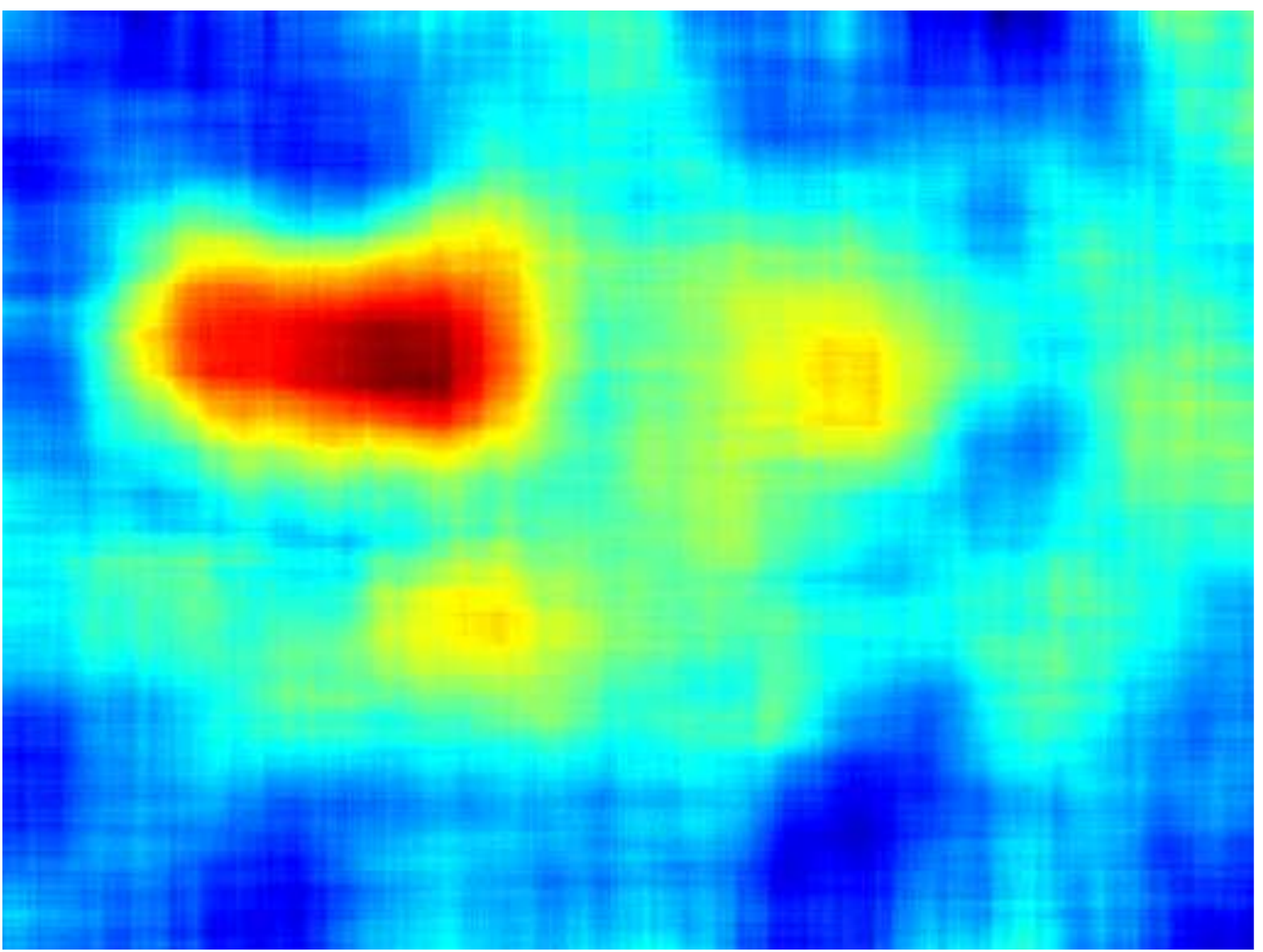}
\fbox{\includegraphics[width=0.11\textwidth]{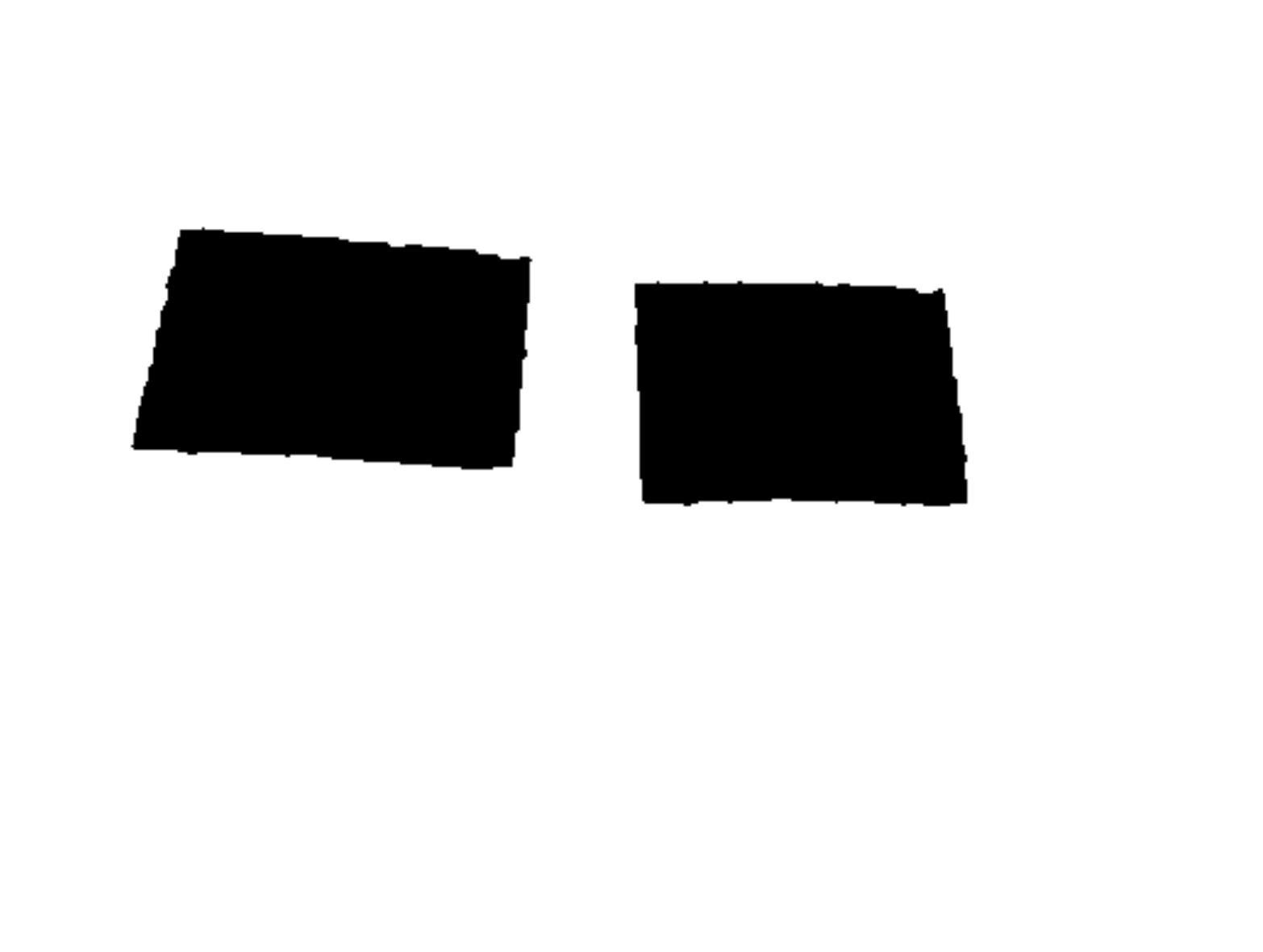}}
\includegraphics[width=0.11\textwidth]{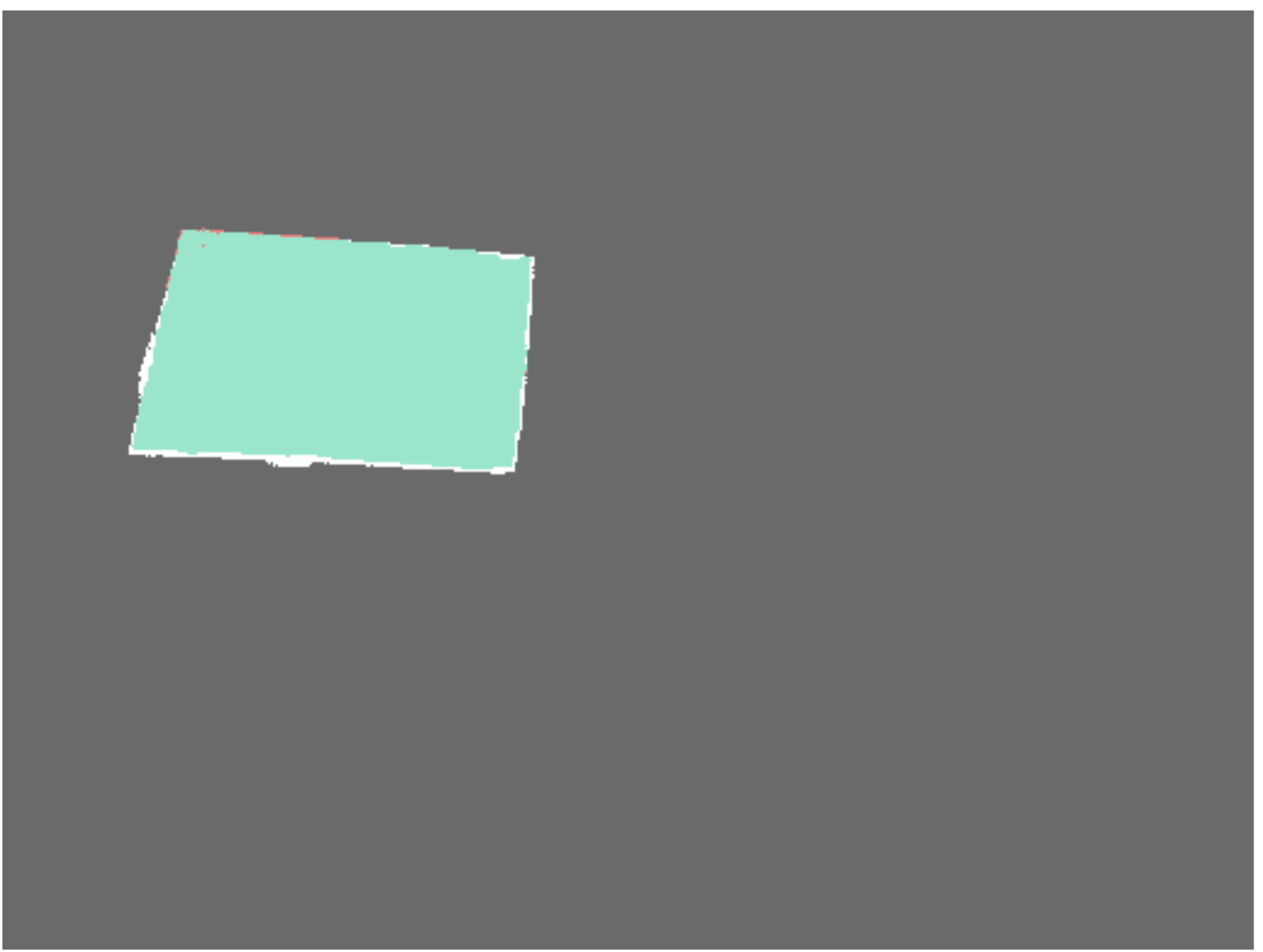}
\caption{A training fake image, its correlation map, its PatchMatch-based map, and the final color-coded mask.} \label{fig:copymoves}
\end{figure}

\section{Splicing localization by local descriptors}

The techniques described above work only on a fraction of all the images,
those with copy-moves forgeries and those for which the PRNU pattern could be estimated.
To integrate this information we propose a novel algorithm effective also on splicings,
namely, objects copied from different images.
In particular,
given the good performance obtained in forgery detection by the local descriptors proposed in \cite{CGV_phase1.13}
we have implemented the same procedure on a sliding-window basis.
The algorithm performs a classification step for each block,
followed by an aggregation phase driven by a suitable reliability measure,
required to merge all data available at a given pixel.

In order to perform classification
a feature extraction process is required with a successive training of a SVM classifier with linear kernel.
Features are computed on 10000 $128\times 128$-pixel blocks, 5000 pristine and 5000 fake, extracted by the training images.
More precisely,
in view of the subsequent integration with a reliable copy-move detector,
we focus on performance for splicings, and train the classifier only on the 144 spliced images found in the training set.
Note that, in this context, a fake block is not a block drawn entirely from a splicing,
but rather a boundary block, since relevant information to discover a forgery is hidden in the transition area.
More precisely,
we label as fake only the blocks which, according to the ground truth, comprise from 20\% to 80\% forged pixels.
The high-pass filter to compute the descriptor is the best one found in phase 1 for detection, a 3rd order linear filter \cite{CGV_phase1.13}.

The image under test is analyzed in a sliding-window modality,
with partially overlapping $128\times 128$-pixels blocks and a 16-pixel step.
For each block we computed the distance of the corresponding feature vector from the SVM hyperplane,
the larger the distance, the more reliable the result.
By aggregating all these values for each pixel we obtain an index related to the probability that the pixel has been tampered,
named SDH (Sum of Distances from the Hyperplane).
The final binary map is obtained by thresolding this index.
An empirical analysis on the training set suggested a threshold equal to 0.25*max(SDH).
Fig.5 shows some sample results.

\begin{figure}[t]
\centering
\includegraphics[width=0.15\textwidth]{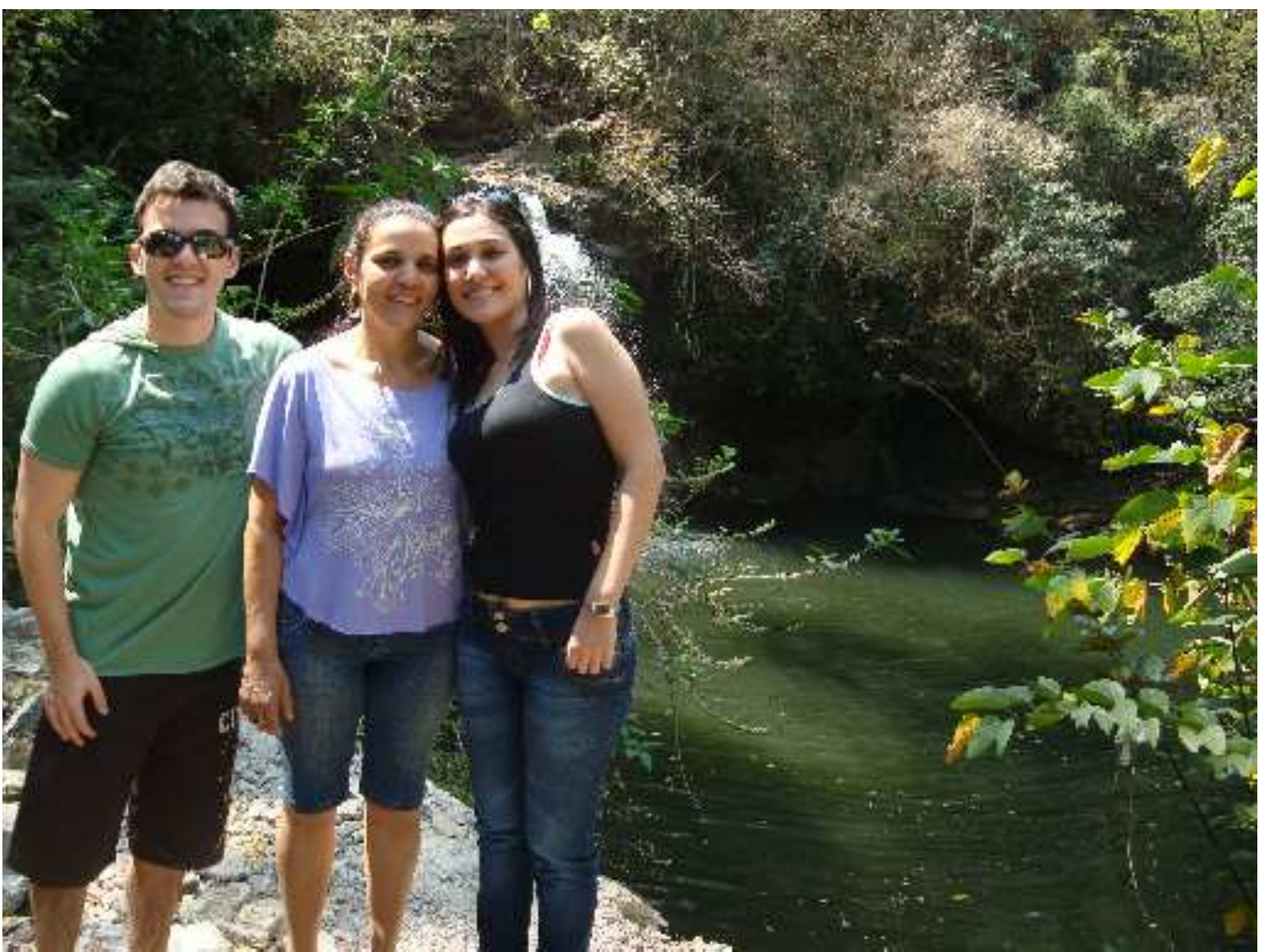}
\includegraphics[width=0.15\textwidth]{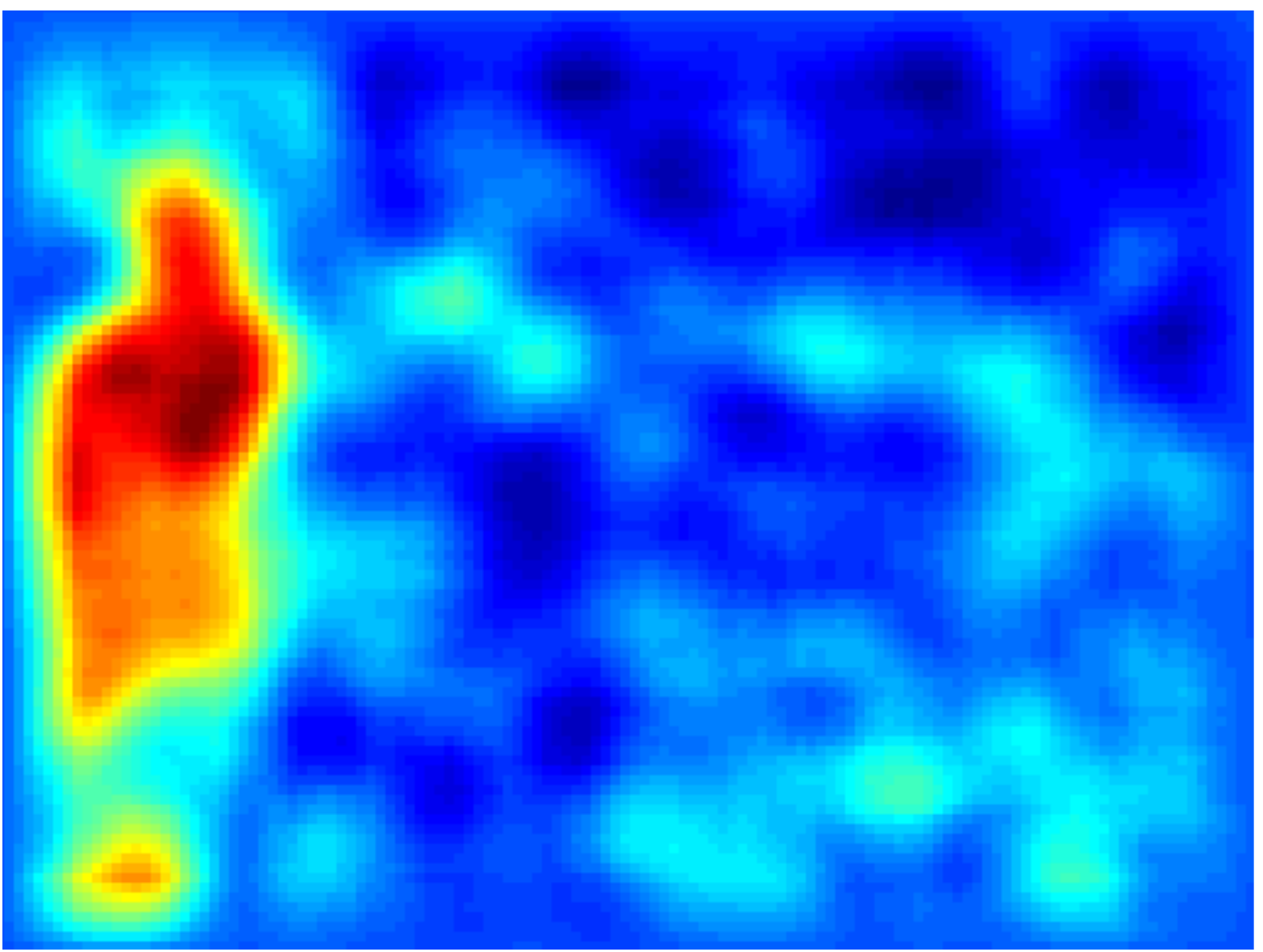}
\includegraphics[width=0.15\textwidth]{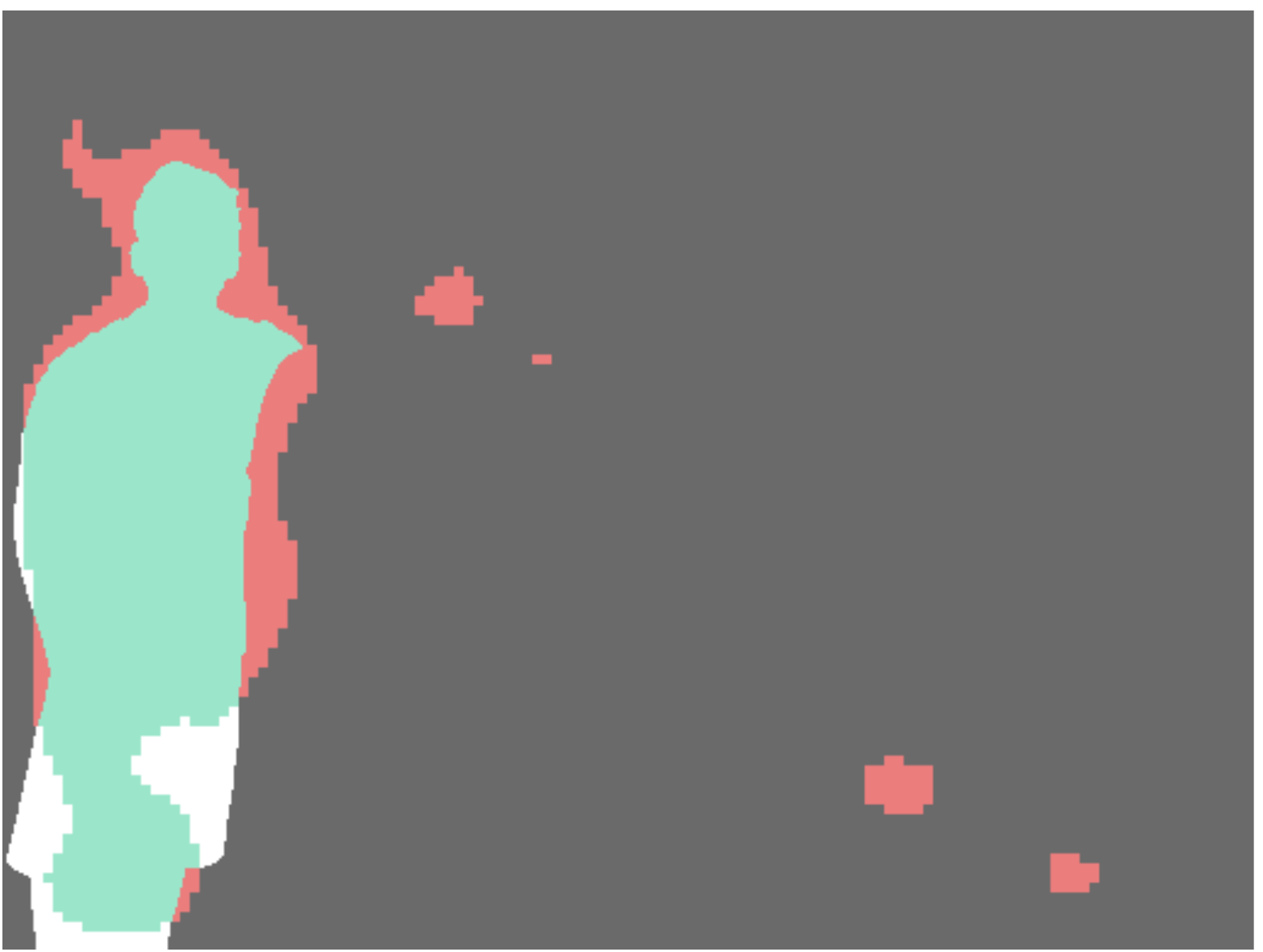}

\vspace{3mm}
\includegraphics[width=0.15\textwidth]{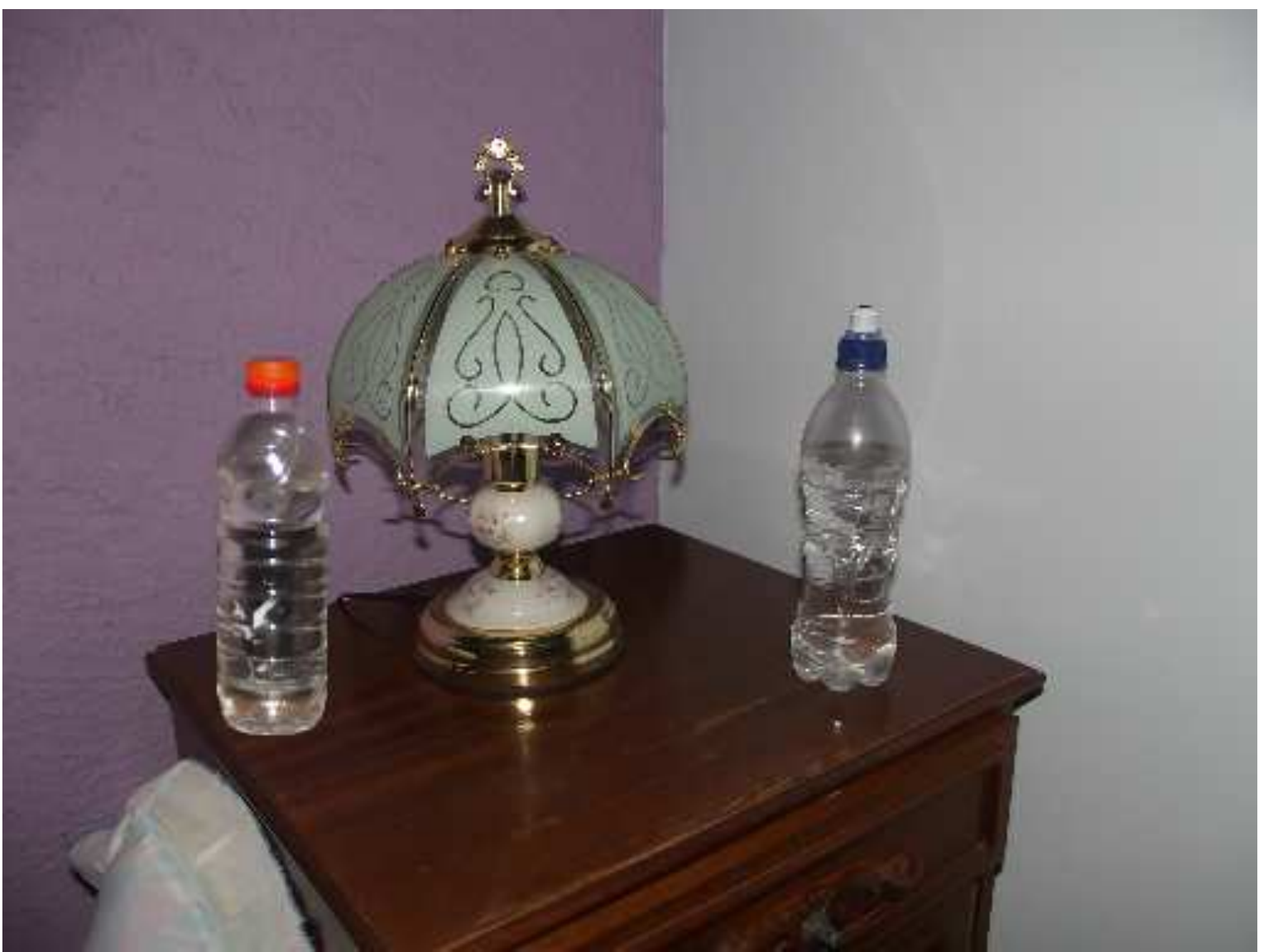}
\includegraphics[width=0.15\textwidth]{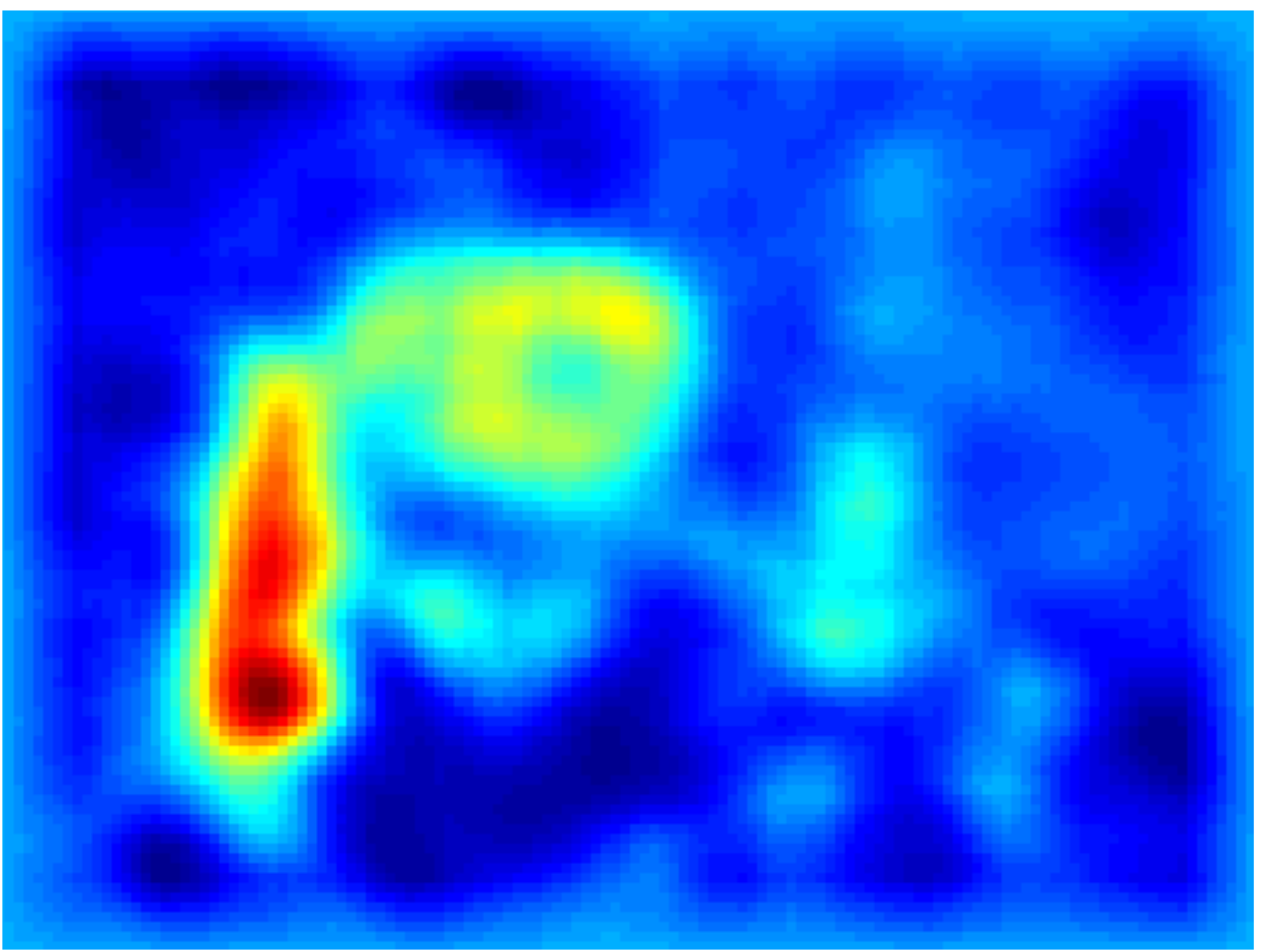}
\includegraphics[width=0.15\textwidth]{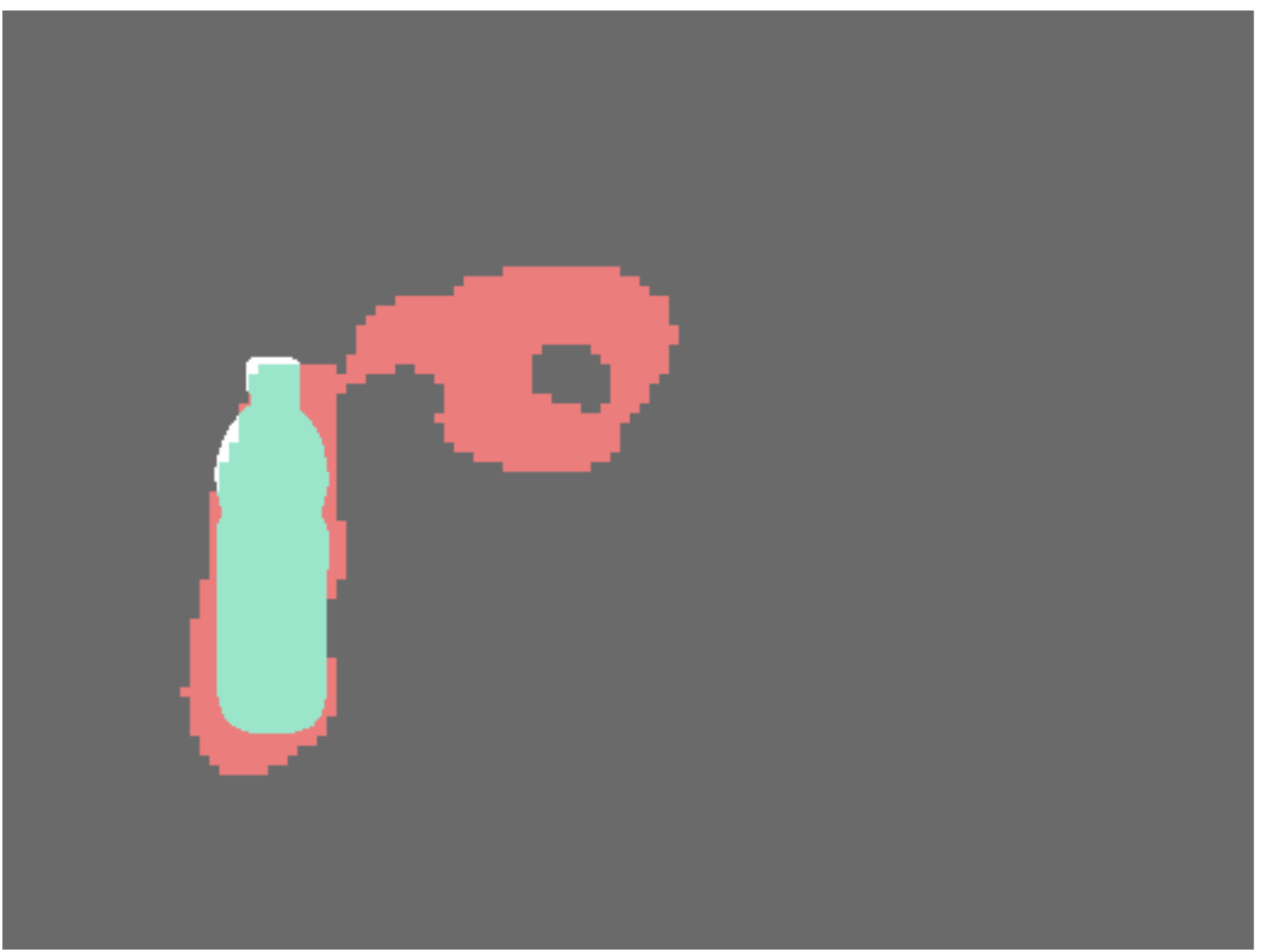}

\caption{Two training fake images, their SDH map and the color coded detection mask.} \label{fig:copymoves}
\end{figure}

\begin{figure}[b]
\centering
\begin{minipage}[c]{.99\linewidth} \centerline{\epsfig{figure=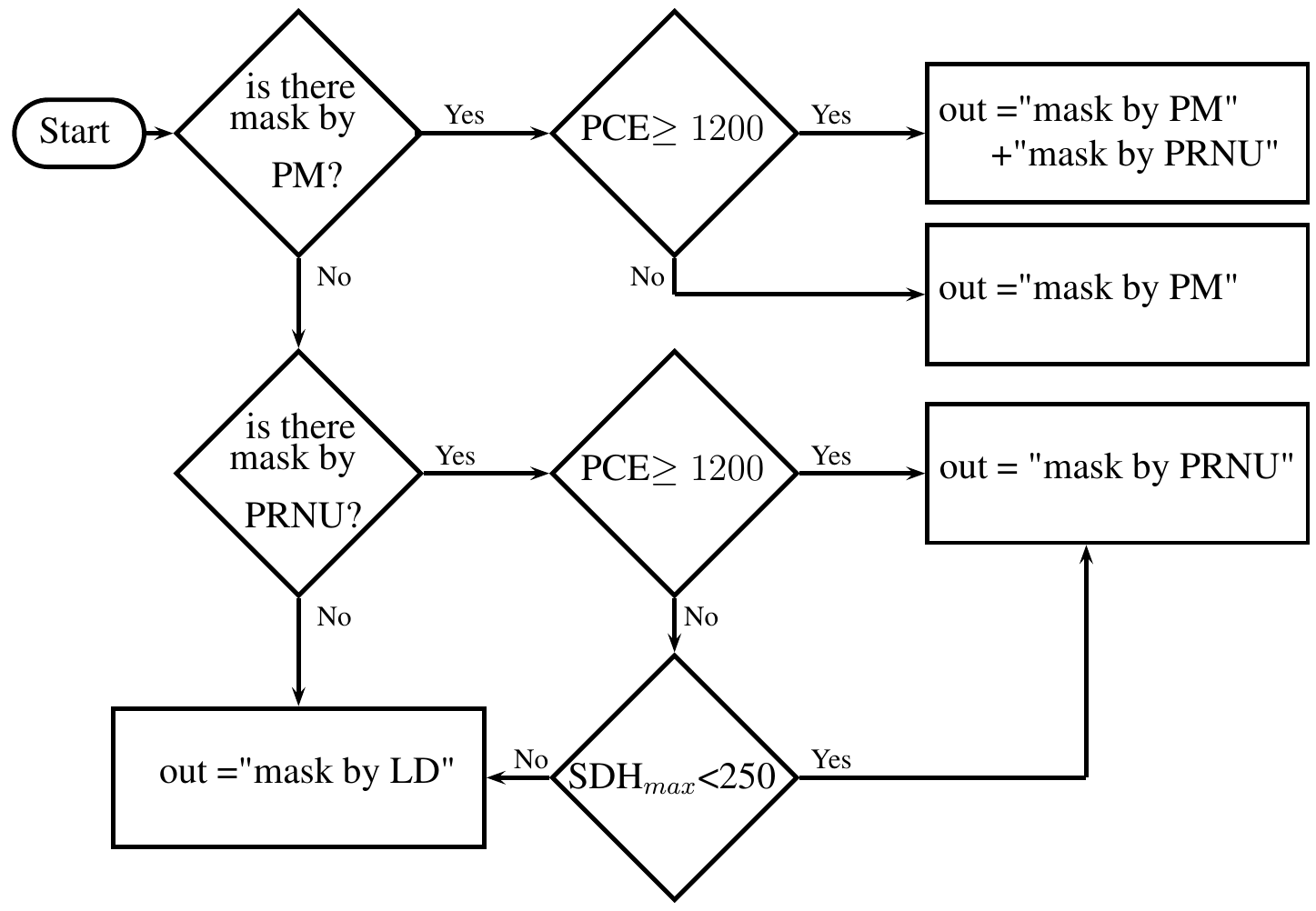, width=8cm}} \end{minipage}
\caption{Flow chart of the combination strategy.}
\label{fig:clusters}
\end{figure}

\section{Combination strategy}

The flow-chart of Fig.6 describes our fusion strategy.
A general guideline was to keep into great account all information about reliability.
In particular,
since F-measure results computed on the training set made very clear the superior reliability of the PatchMatch-based detector,
we use only its map when available, and integrate it with the PRNU-based map only when the latter is itself extremely reliable (PCE$>$1200).
Then when no copy-move is detected, we trust, in decreasing order, the PRNU-based map and the Local Detector map.
It is worth underlining that the latter map,
although less reliable than the previous two, is always available, and hence allows us to make a decision on all the test images.
On the training set, this strategy provided an average F-measure equal to 0.4153,
while on the test set we obtained the best result of phase 2 of the Challenge with 0.4072.
Four sample results on the test set are shown in Fig.7.

\begin{figure}[t]
\centering
\includegraphics[width=0.11\textwidth]{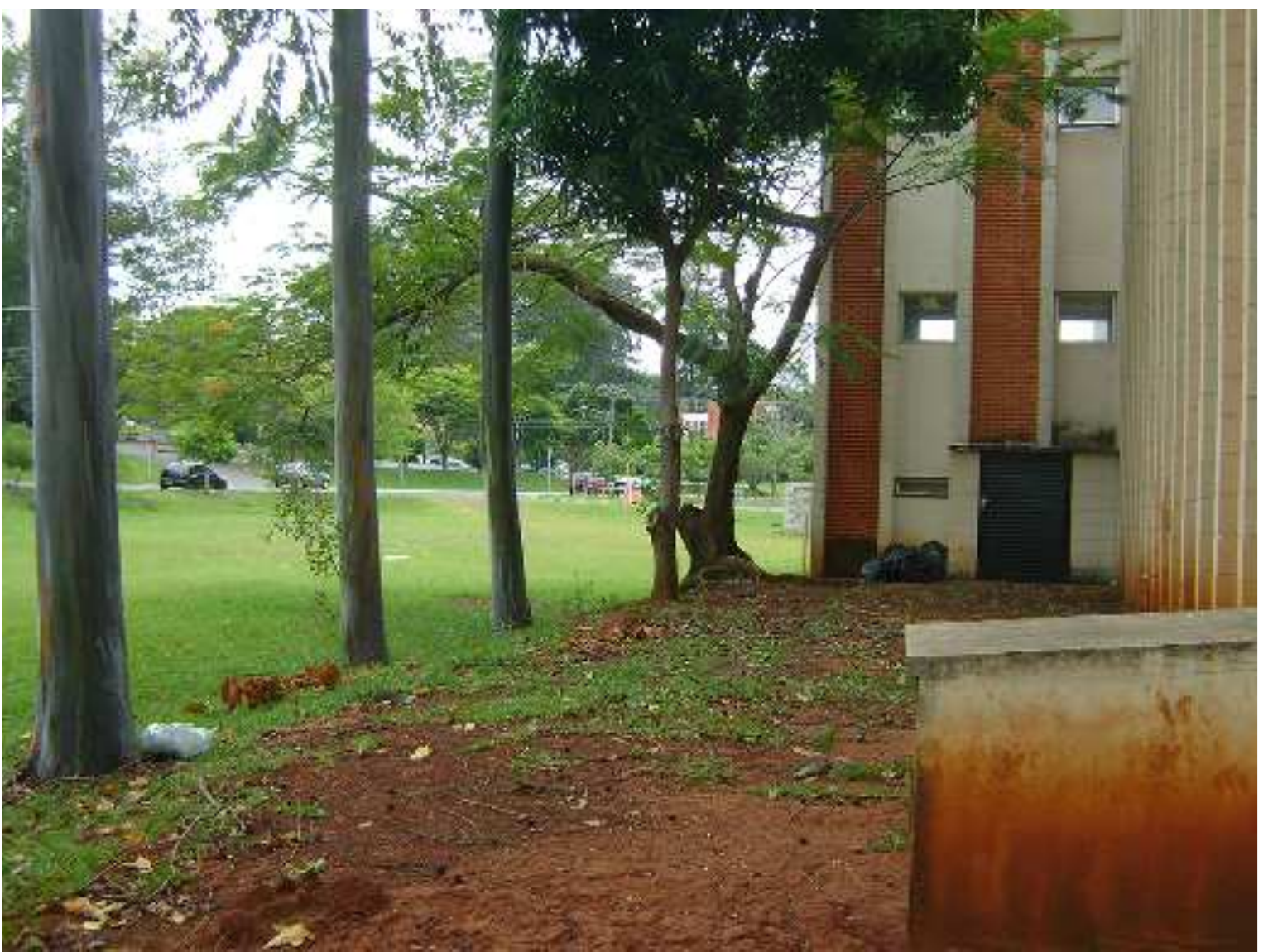}
\fbox{\includegraphics[width=0.11\textwidth]{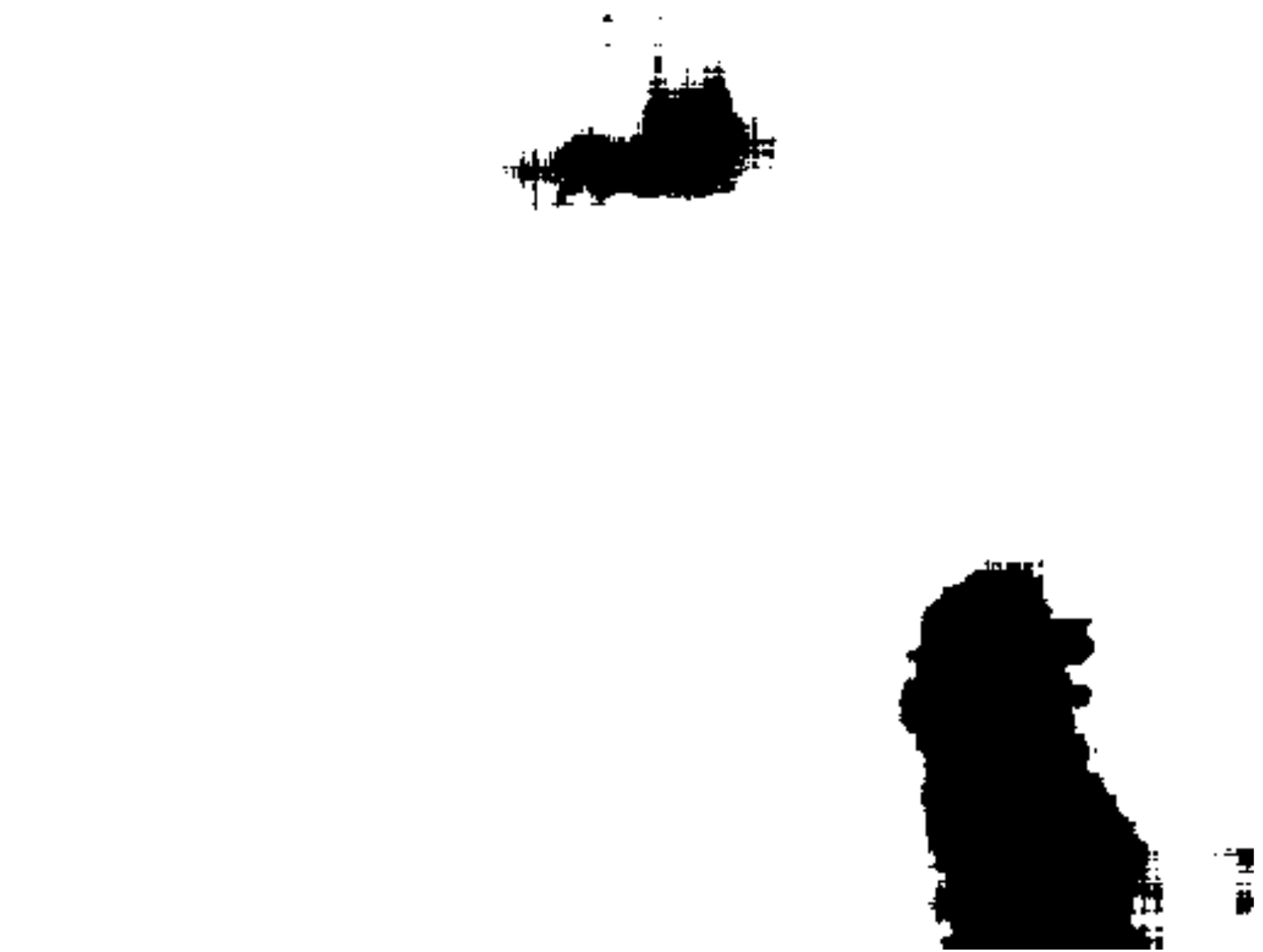}}
\includegraphics[width=0.11\textwidth]{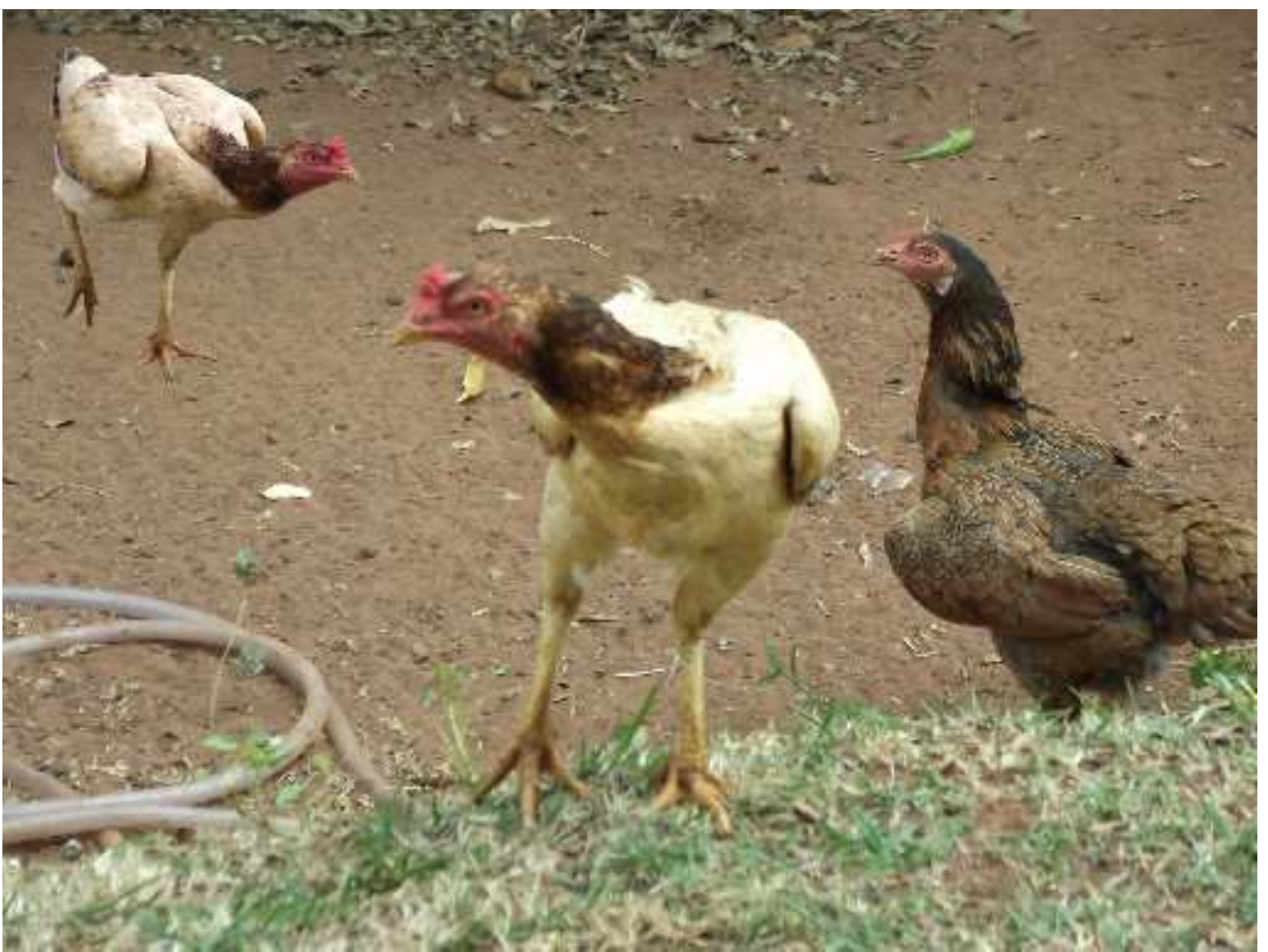}
\fbox{\includegraphics[width=0.11\textwidth]{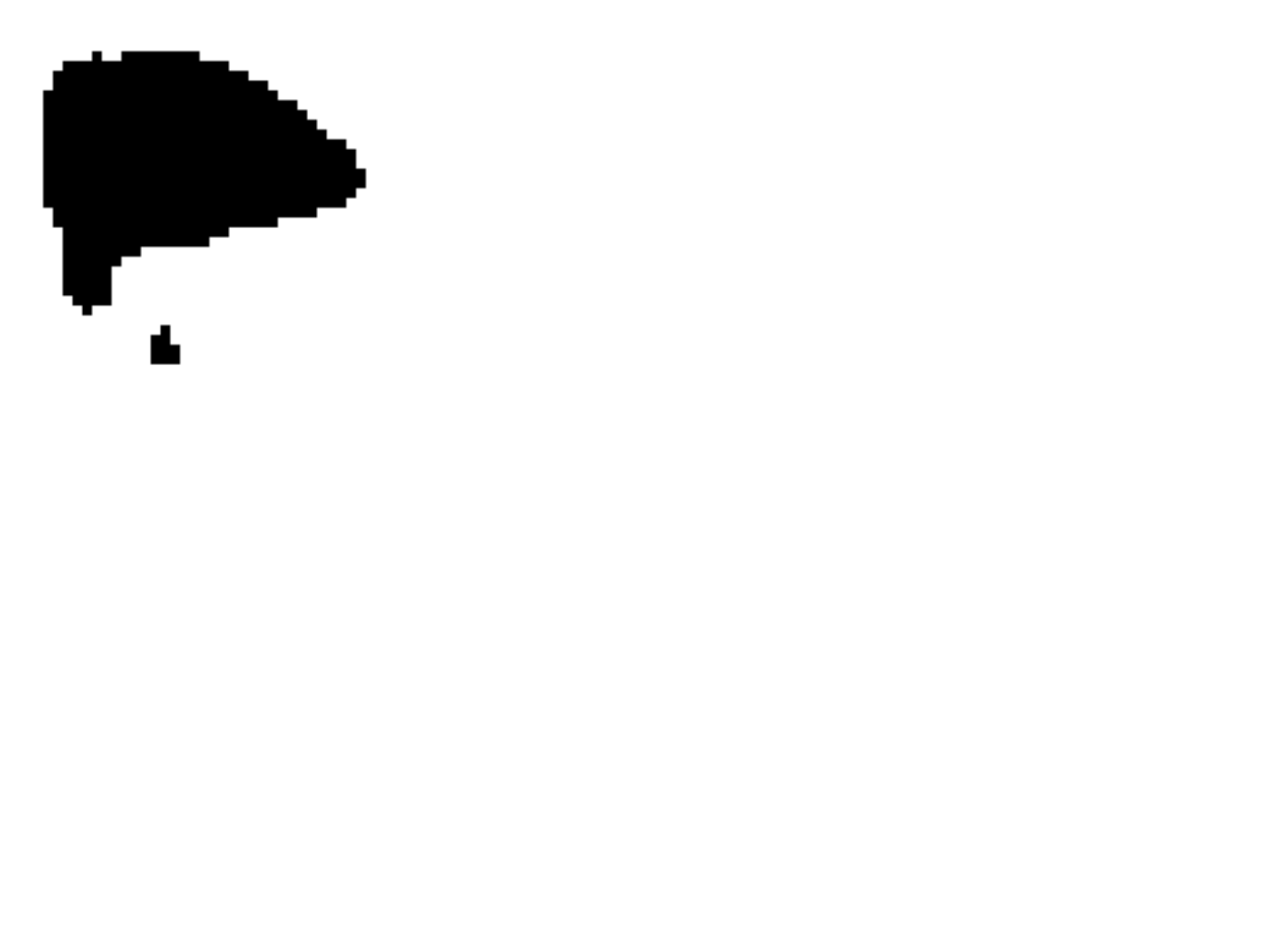}}

\vspace{3mm}
\includegraphics[width=0.11\textwidth]{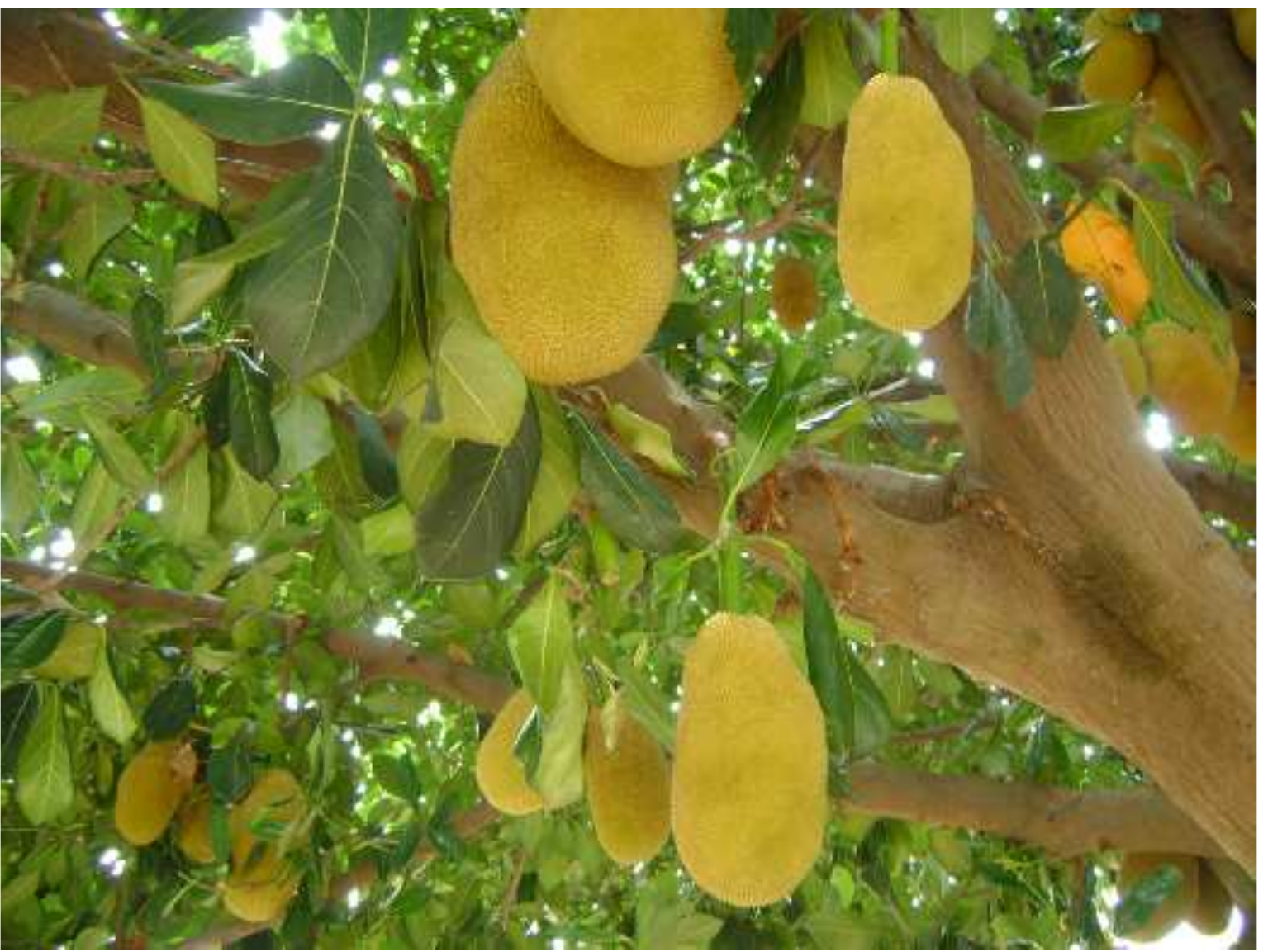}
\fbox{\includegraphics[width=0.11\textwidth]{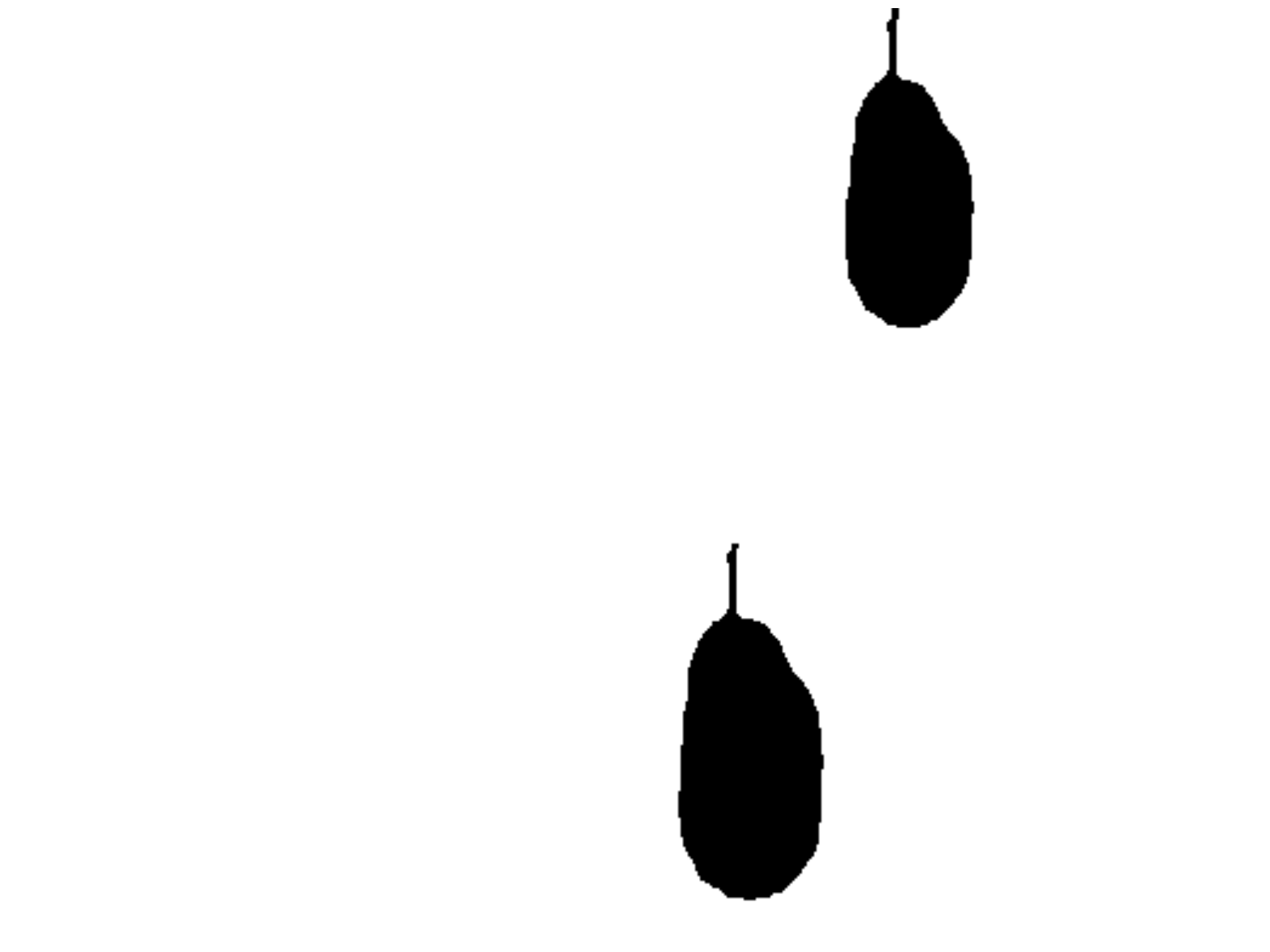}}
\includegraphics[width=0.11\textwidth]{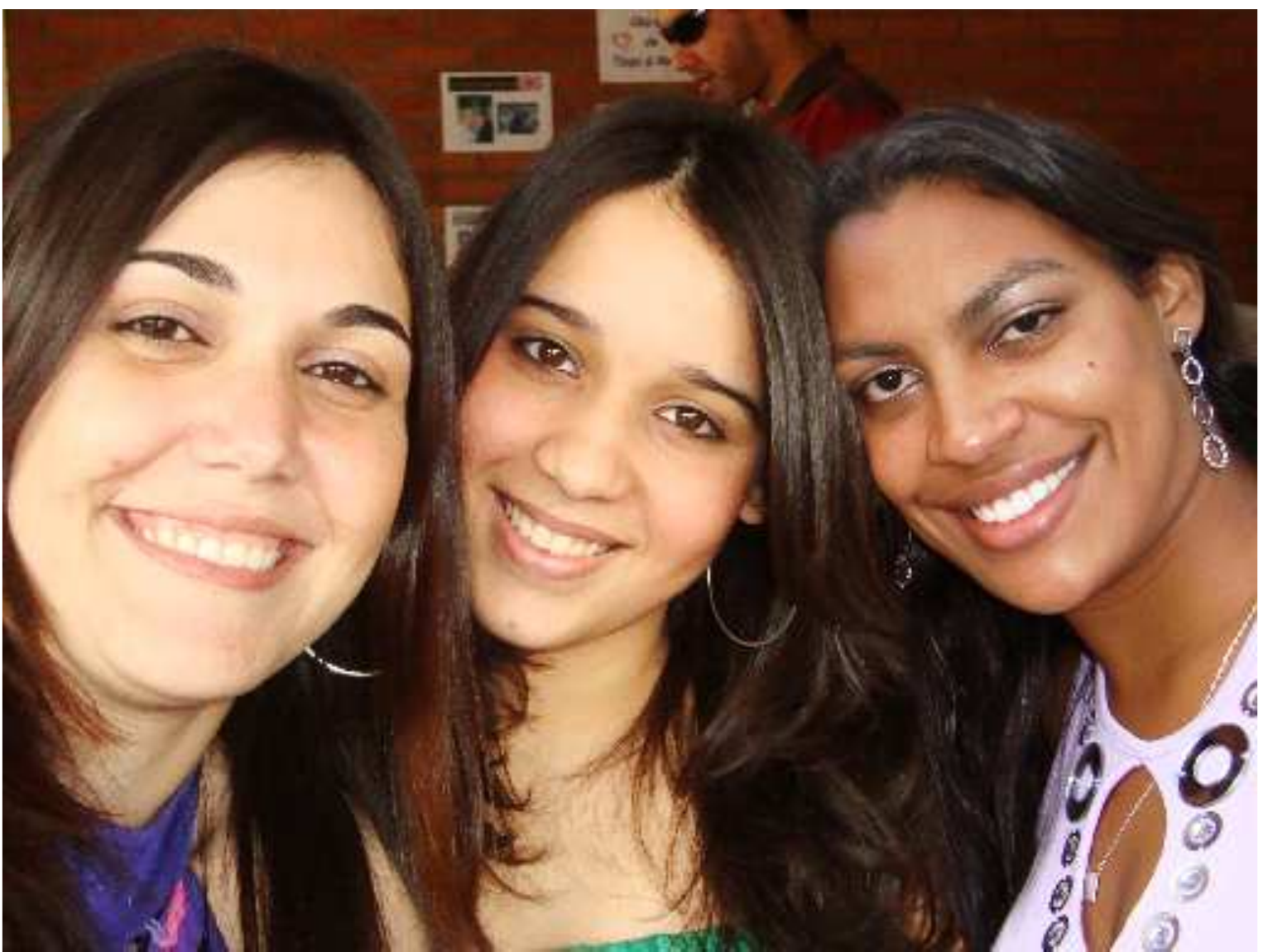}
\fbox{\includegraphics[width=0.11\textwidth]{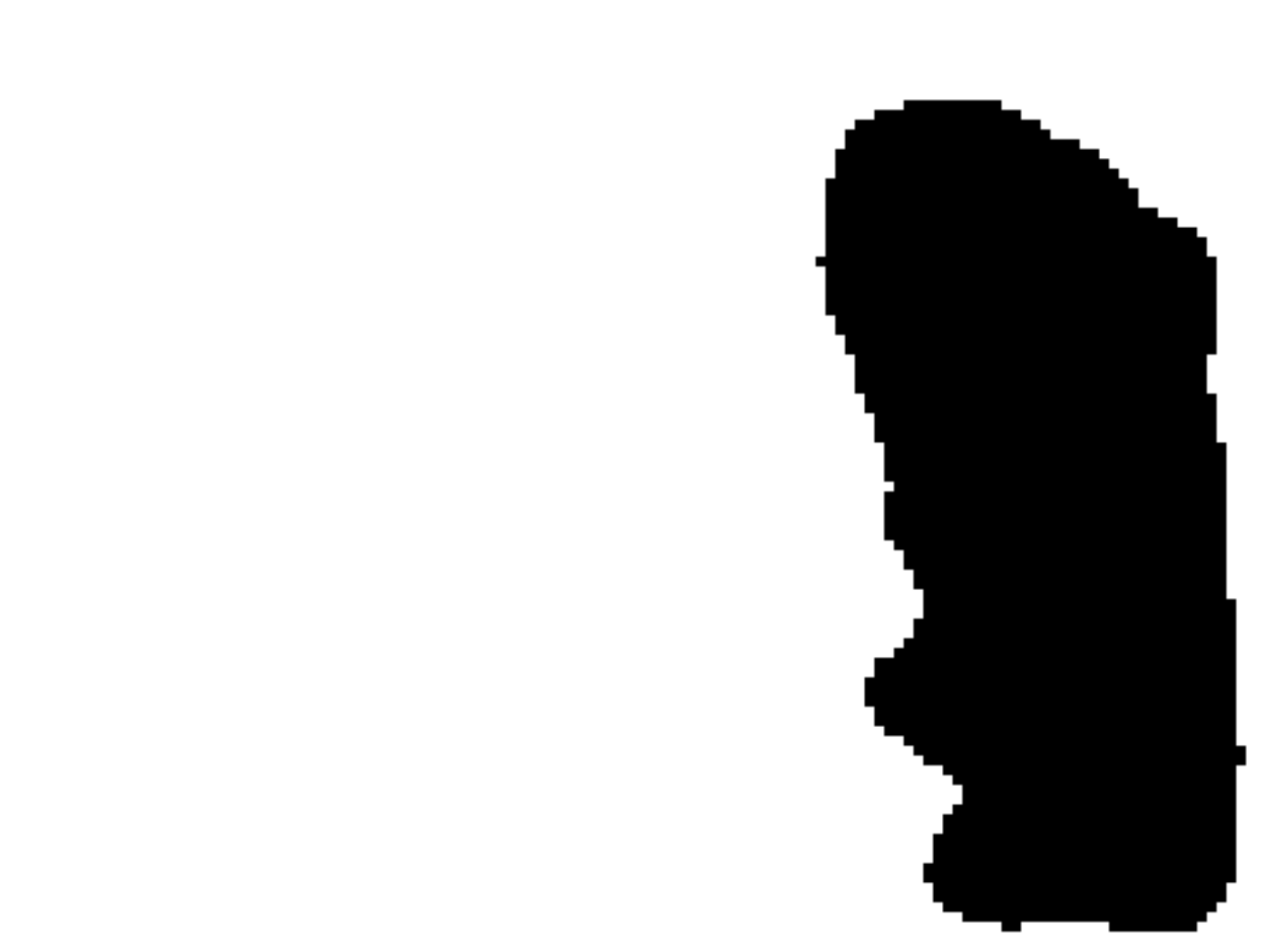}}

\caption{Four images from the test set and their output masks.} \label{fig:copymoves}
\end{figure}

This work confirms that no single tool is sufficient to deal with the diversity of possible image manipulations.
Although we obtained encouraging results, there is ample space for further improvements.
The PRNU-based technique, for example, is not able to detect very small forgeries, and gives too many false alarms in the absence of a predictor.
The PatchMatch-based technique also needs improvements to reduce the false alarm rates.
The LD-based technique is at an embryonal stage and a deep analysis is required to optimize it.
Finally, the information fusion is also rather naive, and a smarter fusion rule could be devised as done for example in \cite{DGSV_iciap.13}.

\end{document}